\documentclass[pdflatex,sn-mathphys-num]{sn-jnl}

\usepackage{graphicx}%
\usepackage{multirow}%
\usepackage{amsmath,amssymb,amsfonts}%
\usepackage{amsthm}%
\usepackage{mathrsfs}%
\usepackage[title]{appendix}%
\usepackage{xcolor}%
\usepackage{textcomp}%
\usepackage{manyfoot}%
\usepackage{booktabs}%
\usepackage[ruled,vlined]{algorithm2e}
\SetKwInput{KwInput}{Input}       
\SetKwInput{KwInit}{Initialize}   
\usepackage{algpseudocode}

\usepackage{listings}%
\usepackage{enumitem}

\newenvironment{unnumberedlemma}[1][]{%
    \par\vspace{1ex}\noindent\textbf{Lemma }\hspace{0.5em}\itshape
}{\par\vspace{1ex}}

\newenvironment{unnumberedcorollary}[1][]{%
    \par\vspace{1ex}\noindent\textbf{Corollary }\hspace{0.5em}\itshape
}{\par\vspace{1ex}}

\theoremstyle{thmstyleone}%
\newtheorem{theorem}{Theorem}
\newtheorem{lemma}[theorem]{Lemma} 
\newtheorem{corollary}[theorem]{Corollary}
\theoremstyle{thmstyletwo}%
\newtheorem{remark}{Remark}%

\theoremstyle{thmstylethree}%

\begin{document}



\title{Prequential posteriors}


\author[1]{\fnm{Shreya} \sur{Sinha Roy}}\email{shreya.sinha-roy@warwick.ac.uk}

\author[1]{\fnm{Richard G.} \sur{Everitt}}\email{richard.everitt@warwick.ac.uk}

\author[1,2]{\fnm{Christian P.} \sur{Robert}}\email{c.a.m.robert@warwick.ac.uk}

\author[1]{\fnm{Ritabrata} \sur{Dutta}}\email{ritabrata.dutta@warwick.ac.uk}

\affil[1]{\orgdiv{Department of Statistics}, \orgname{University of Warwick}, \orgaddress{ \city{Coventry}, \postcode{CV4 7AL}, \country{United Kingdom}}}

\affil[2]{\orgdiv{CEREMADE}, \orgname{Université Paris Dauphine PSL}, \orgaddress{\street{16 Bd Lannes}, \city{Paris}, \postcode{75116}, \country{France}}}


\abstract{
    Data assimilation is a fundamental task in updating forecasting models upon observing new data, with applications ranging from weather prediction to online reinforcement learning. Deep generative forecasting models (DGFMs) have shown excellent performance in these areas, but assimilating data into such models is challenging due to their intractable likelihood functions. This limitation restricts the use of standard Bayesian data assimilation methodologies for DGFMs. To overcome this, we introduce \textit{prequential posteriors}, based upon a predictive-sequential (prequential) loss function; an approach naturally suited for temporally dependent data which is the focus of forecasting tasks. Since the true data-generating process often lies outside the assumed model class, we adopt an alternative notion of consistency and prove that, under mild conditions, both the prequential loss minimizer and the prequential posterior concentrate around parameters with optimal predictive performance. For scalable inference, we employ easily parallelizable wastefree sequential Monte Carlo (SMC) samplers with preconditioned gradient-based kernels, enabling efficient exploration of high-dimensional parameter spaces such as those in DGFMs. We validate our method on both a synthetic multi-dimensional time series and a real-world meteorological dataset; highlighting its practical utility for data assimilation for complex dynamical systems.
}


\maketitle

\section{Introduction}

Data assimilation \citep{vetra2018state} refers to the process of combining observational data with a computational model to estimate the evolving state of a dynamical system. It plays a crucial role in domains such as weather forecasting, meteorology, climate modeling, sequential decision-making, and online reinforcement learning. A standard approach for modeling time series data, keeping data assimilation feasible, is through {state space models} (SSMs). In SSMs, the latent state evolves according to a dynamic model, coupled with an observational error model. Under the classical assumption of Gaussian errors and linear dynamics, inference can be performed via the Kalman filter. However, this setting is insufficient for nonlinear and complex dynamics, leading to extensions such as the extended Kalman filter (EKF) \citep{ribeiro2004kalman} and the ensemble Kalman filter (EnKF) \cite{evensen2003ensemble}. On the other hand the physics-based partial differential equation models \citep{fisher2009data} are the work horse of the weather community, but data-assimilation for them tend to be computationally expensive. In the weather forecasting community, variational approaches such as 3D-Var and 4D-Var \citep{courtier1998ecmwf, andersson1994use} have been proposed as optimization-based filtering and smoothing algorithms, but their computational cost is often prohibitive.

Recent years have seen a surge of DNN-based models \citep{arcucci2021deep, qu2024deep} providing fast, low-cost simulations and capturing complex nonlinear dynamics, as alternatives to numerical weather prediction (NWP) models. Prominent deterministic forecasting systems include Pangu-Weather \citep{bi2022panguweather3dhighresolutionmodel}, FourCastNet \citep{pathak2022fourcastnetglobaldatadrivenhighresolution}, GraphCast \citep{lam2023graphcastlearningskillfulmediumrange} and ClimaX \citep{pmlr-v202-nguyen23a}. These models, typically trained with mean squared error (MSE) losses or their variants, achieve impressive accuracy but fail to provide uncertainty quantification. To address this limitation, deep generative forecasting models (DGFM) such as GenCast \citep{price2024gencastdiffusionbasedensembleforecasting} employ generative modeling for uncertainty-aware forecasts. While adversarial training is common in generative modeling, \citet{pacchiardi2024probabilistic} demonstrate that proper scoring rules (SRs) \citep{gneiting2007strictly, waghmare2025proper} can yield calibrated probabilistic forecasts without adversarial objectives from DGFMs. This has also been adapted for training of DGFMs used for probabilistic forecasting by the European centre for midrange weather forecasting (ECMWF) \citep{lang2024aifscrpsensembleforecastingusing}.

A central challenge in implementing Bayesian data assimilation for DGFMs is the lack of a tractable likelihood function, which is fundamental to classical Bayesian inference. Likelihood-free approaches such as approximate Bayesian computation (ABC) \citep{lintusaari2017fundamentals, bernton2019approximate} and synthetic likelihood \citep{price2018bayesian, an2020robust, thomas2020likelihood} exploit model simulation to define a posterior, but face scalability issues. Alternatively, generalized posteriors based on loss function \citep{bissiri2016general, jewson2018principles, knoblauch2022optimization} provide a more flexible framework for probabilistic inference with generative models. Recent work by \citet{pacchiardi2024generalized} extends this framework to deep generative models by defining posteriors over neural network parameters. However, while such loss-based methods are well studied for the independent and identically distributed (IID) data setting, their extension to time-dependent data remains limited; leaving a gap in applying data assimilation for generative models.

In this work, we address this gap by introducing the notion of a \textit{prequential posterior} for data assimilation. Instead of relying on the (often unavailable) likelihood, we replace it with a prequential loss \citep{dawid1991fisherian, dawid1992prequential, dawid1999prequential}, which directly evaluates predictive performance on sequential data. This formulation is especially well suited for probabilistic forecasting, where the objective is not only to achieve accurate predictions but also to quantify predictive uncertainty. However, this raises a fundamental issue: the true data-generating process is rarely contained within the assumed model class. Under such model misspecification, classical consistency results (e.g., convergence of the MLE to the true parameter) no longer apply.

To overcome this, we adopt a different notion of predictive consistency \citep{skouras2000consistency}, which focuses on selecting the parameter that minimizes the one-step-ahead predictive risk rather than recovering a nonexistent “true” parameter. We show that the minimizer of the empirical prequential loss is consistent in this predictive sense and establish a Bernstein–von Mises-type result for the generalized prequential posterior, demonstrating concentration around the optimal predictive parameter.

Sequential Monte Carlo (SMC) methods \citep{del2006sequential} have been widely used for data assimilation across domains such as geosciences, engineering, and machine learning, with applications ranging from earthquake forecasting \citep{werner2011earthquake} to online reinforcement learning \citep{roy2024generalized}. In this work, we implement an easily parallelizable, waste-free SMC algorithm \citep{dau2022waste}, equipped with a gradient-based, preconditioned forward kernel to ensure efficient exploration of high-dimensional parameter spaces. This kernel is inspired by stochastic gradient Markov chain Monte Carlo (SG-MCMC) methods \citep{welling2011bayesian, chen2014stochastic, ding2014bayesian} and adaptive stochastic methods \citep{jones2011adaptive}, where particle transitions are guided by gradient information. Additionally, we incorporate RMSprop-style preconditioning \citep{chen2016bridging} to enhance stability and performance in high-dimensional settings.

Together, these components form a likelihood-free, scalable, and adaptive Bayesian framework for data assimilation, which is applicable for DGFMs. The main contributions of our work are summarized below:

\begin{enumerate}
\item A likelihood-free data assimilation framework for generative forecasting models via prequential posteriors.
\item Theoretical analysis of the prequential posterior, establishing predictive consistency under model misspecification.
\item A scalable and computationally efficient SMC algorithm for training high-dimensional neural network parameters.
\end{enumerate}

The remainder of the paper is structured as follows. Section~\ref{ChapPPsec: gen bayes} introduces the posterior formulation and presents an asymptotic analysis in Section~\ref{ChapPPsubsec: consistency} using a different notion of consistency for a generic loss function. In Section~\ref{ChapPPsubsec: sr}, we introduce scoring rules as a special class of loss functions, which we use to define the prequential posterior. Our proposed sequential inference method via SMC is described in Section~\ref{ChapPPsec: sampling scheme}. Section~\ref{ChapPPsec: sim study} presents our simulation studies, where 
we first implement our proposed method on a well specified stationary time series. Next, we compare our data assimilation method with a misspecified ensemble Kalman filter (EnKF) for a time series model called Lorenz 96 \citep{lorenz1996predictability} and demonstrate its superior long-run predictive performance. We also apply the approach to a real-world meteorological forecasting task using the WeatherBench dataset~\citep{rasp2020weatherbench}. The code for reproducing these results is available \href{https://github.com/ShreyaSR999/Prequential-posteriors}{here}. Finally, we conclude in Section~\ref{ChapPPsec: conclusion}, where we summarize our findings and discuss potential directions for future research.

\section{Prequential posterior for data-assimilation of DGFMs} \label{ChapPPsec: gen bayes}

Consider a discrete-time stochastic process \((Y_1, Y_2, \ldots) \), where each \(Y_t \in \mathcal{Y} \subseteq \mathbb{R}^l\). We assume that these random variables are not independent; rather, they exhibit temporal dependencies.
Suppose that the true underlying process generating the observations is denoted by \( P \). Given that the observations exhibit temporal dependence, it is reasonable to assume that the distribution of the observation at time \( t \) depends on the entire history of observations up to time \( t-1 \). Formally, we express this by stating that
\[
Y_t \sim P_t, \quad \text{where } P_t = P(\cdot \mid Y_{1:t-1}),
\]
where we use the notation \( Y_{i:j} \) to denote the sequence of observations \( Y_i, Y_{i+1}, \ldots, Y_j \).


\paragraph{Deep generative forecasting models (DGFMs)}
Probabilistic forecasting seeks not only accurate predictions of future 
observations but also reliable uncertainty estimates. A common way to achieve 
this is via a \textit{simulator model} $Q^{\theta}$ that generates forecasts 
at time $t$ from past data $Y_{1:t-1}$:
\[
\hat{Y}_t \sim Q^{\theta}_t, \quad \theta \in \Theta \subseteq \mathbb{R}^p.
\]
If the true data-generating distribution $P$ lies in the model class 
$\mathcal{Q} = \{Q^{\theta} : \theta \in \Theta\}$, then the model is 
\textit{well-specified}, with some $\theta^*$ satisfying $Q^{\theta^*} = P$.
Deep generative models provide a natural choice for $Q^{\theta}$, as they can 
capture nonlinear, high-dimensional dynamics while producing samples rather 
than point predictions. In this case, the forecast is generated by pushing 
auxiliary randomness $W$ through a neural network:
\[
\hat{Y}_t = Q_t^{\theta}(Y_{1:t-1}, W),
\]
where $W$ is independent noise and $\theta$ are the network parameters. We 
refer to this class of models as \textit{deep generative forecasting models (DGFMs)}, 
which combine the expressiveness of deep learning with uncertainty-aware 
simulation for forecasting tasks \citep{price2024gencastdiffusionbasedensembleforecasting, pacchiardi2024probabilistic}.
\smallskip



Bayesian inference on these model parameters typically requires a tractable likelihood function, which poses a significant challenge in the context of generative models. \cite{pacchiardi2024generalized} provided a likelihood-free framework for Bayesian inference for deep generative models. 
We extend this work to settings with \textit{temporal dependencies} in the observations, drawing inspiration from the prequential framework proposed by \citet{dawid1991fisherian}, \cite{dawid1984present}, \cite{dawid1992prequential}, \cite{dawid1999prequential}.

\subsection{Prequential posteriors}

Let the sequence of random variables \( Y = (Y_1, Y_2, \ldots) \) be defined on a complete filtered probability space \( (\Omega, \mathcal{F},( \mathcal{F}_t), P) \), where each \( Y_t \) is \( \mathcal{F}_t \)-measurable and the filtration \( \mathcal{F}_t \) is the \(\sigma\)-algebra generated by  \( Y^t = (Y_1, Y_2, \ldots, Y_t) \).

\begin{figure}[ht]
    \centering
    \includegraphics[width=0.5\linewidth]{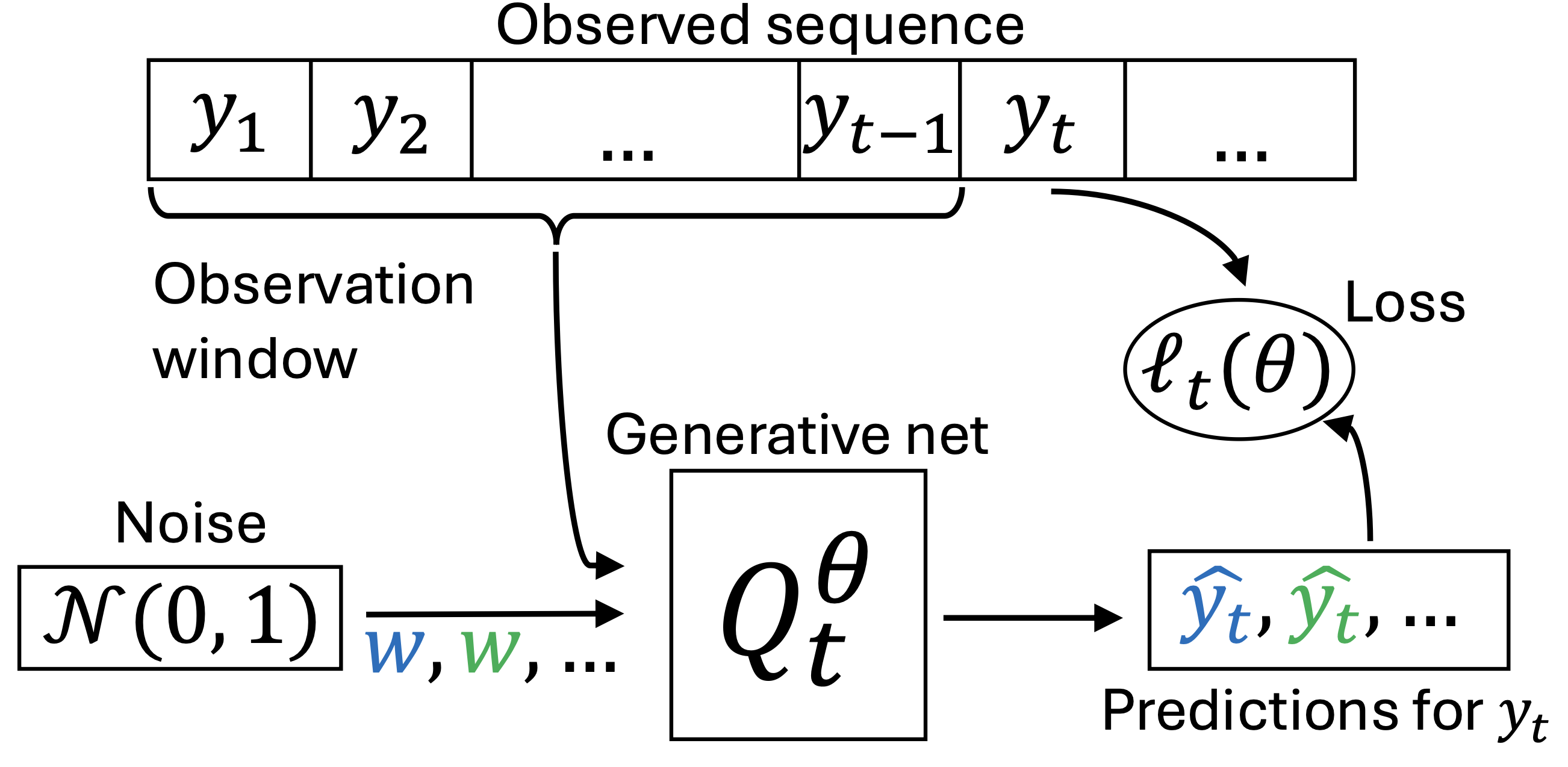}
    \caption{Illustration of the loss calculation. The loss function 
    $\ell_T^{\theta}$ evaluates forecasts generated by the conditional 
    generative network $Q_t^{\theta}$, which predicts $Y_t$ given the past 
    observations $Y_{1:t-1}$; with respect to the observed $y_t$.}
    \label{ChapPPfig: loss diagram}
\end{figure}

We assign a time-indexed, \( \mathcal{F}_t \)-measurable loss function \( \ell_t(\theta) \) to the simulator model \( Q^{\theta}_t \), which produces a forecast \( \hat{Y}_t \) at time \( t \). This loss function evaluates the discrepancy between the forecast \( \hat{Y}_t \) and the actual observation \( Y_t \), conditioned on the past observations $Y_{1:t-1}$ as shown in Figure~\ref{ChapPPfig: loss diagram}.

Given an observed sequence \( Y^T = (Y_1, Y_2, \ldots, Y_T) \), we define the \textit{prequential (predictive sequential)} loss for the simulator model \( Q^{\theta} \) as
\[
L_T(\theta) = \sum_{t=1}^{T} \ell_t(\theta). 
\]
Thus, \( L_T(\theta) \) is a \( \mathcal{F}_T \)-measurable loss which assesses the predictive performance of the model $Q^{\theta}$ with respect to the sequence of observations $Y^T$.


Let us assume a prior distribution \( \pi(\cdot) \), which is a probability measure with respect to the measurable space $(\Theta, \sigma({\Theta}))$. Upon observing a realization \( y^T \) of the sequence \( Y^T \), we define the prequential posterior (PP) 
as,

\begin{equation}\label{ChapPPeqn: preq-posterior}
    \pi_{P}(\theta \mid y^T) \propto \pi(\theta) \exp(- \gamma L_T(\theta)),
\end{equation}
where $\gamma$ is a free parameter which controls how much the cumulative loss influences the posterior relative to the prior distribution. By taking $\gamma=1$ and the loss function to be the negative log likelihood function of a time series model with known likelihood functions, we recover the standard Bayesian posterior distribution for the time series model from the prequential posterior. 

\subsection{Consistency of the prequential posteriors}\label{ChapPPsubsec: consistency}
Before delving into the asymptotic properties of the prequential score and the posterior it induces, let us first consider the case where the observations are assumed to be IID and suppose the model likelihood is known. The uniform law of large numbers (ULLN) 
leads to the asymptotic consistency of the maximum likelihood estimator to the true parameter ($\theta^*$).

However, this classical result breaks down when strong dependence exists between successive observations. In such cases, the ULLN does not apply naturally. The consistency of extremum estimators, those that minimize the prequential loss, has instead been established under asymptotic mixing conditions on the data-generating process. Loosely speaking, these conditions imply that that as $m \to \infty$, $Y_{t-m}$ and $Y_t$ become independent \citep{pacchiardi2024probabilistic}. This covers settings such as fixed-order Markov processes, but does not extend to models with strong or long-range temporal dependencies.

Nevertheless, a fundamental challenge arises when the true data-generating process does not lie within the assumed model class: how should we define consistency in such cases? To illustrate this, we draw on an example from \cite{skouras2000consistency}, which highlights the need for an alternative notion of consistency.

\paragraph{Example from \cite{skouras2000consistency}} Suppose that $(X_t)$ is a sequence of IID random variables distributed as $\mathcal{N}(0, 1)$. Define the observed process as $Y_t = X_0 + X_t$. Consider a model $\mathbb{E}_{t-1}(Y_t) = \theta$ for some constant $\theta$. We estimate $\theta$ by minimizing the squared error loss:
\[
\bar{L}_T(\theta) = \frac{1}{T} \sum_{t=1}^T (Y_t - \theta)^2,
\]
whose minimizer is
\[
\hat{\theta}_T = \arg \min_{\theta} L_T(\theta) = \frac{1}{T} \sum_{t=1}^T Y_t.
\]
However, the expected loss is given by:
\[
\bar{L}_T^{**}(\theta) = \mathbb{E} \bar{L}_T(\theta) =  \frac{1}{T} \sum_{t=1}^T \mathbb{E}(Y_t - \theta)^2 = \mathbb{E}(Y_t^2) + \theta^2 = 2 + \theta^2,
\]
which is minimized at $\theta^{**} = 0$. On the other hand, we note that $\hat{\theta}_T = X_0 + \frac{1}{T} \sum_{t=1}^T X_t$, which converges almost surely to the observed value $x_0$ of $X_0$ as $T \to \infty$. Therefore, $\hat{\theta}_T \not\to \theta^{**}$, but instead converges to a data-dependent limit.

These limitations motivate us to explore an alternative framework for analyzing the consistency of prequential loss minimizers introduced in \citet{skouras2000consistency} and \citet{dawid2020learnability}. Note that both the true observation \( Y_t \) and the prediction of the model \( \hat{Y}_t \) at time $t$ depend on previous observations \( Y_{1:t-1} \). As a result, different histories of \( Y_{1:t-1} \) can yield different conditional distributions for \( Y_t \). In the case of non-ergodic models, this dependence on past observations may persist indefinitely. Therefore, it is more appropriate to assess predictive performance using conditional expectations rather than marginal ones. Hence, we define the conditional predictive risk as the conditional expected loss at time \( t \) given by:
\[
\ell^*_t(\theta) = \mathbb{E}_{t-1}[\ell_t(\theta)], 
\]
where \( \mathbb{E}_{t-1}(\cdot) = \mathbb{E}(\cdot \mid \mathcal{F}_{t-1}) \) denotes the conditional expectation given the filtration \( \mathcal{F}_{t-1} \), i.e., all observations up to time \( t-1 \).

 The conditionally expected score \( \ell^*_t(\theta) \) thus captures the one-step ahead predictive risk of the model, given the observed history up to time \( t-1 \). We denote the expected prequential loss by,  

\[
L^*_T(\theta) = \sum_{t=1}^T \mathbb{E}_{t-1}[\ell_t(\theta)].   
\]

Returning to the example from \cite{skouras2000consistency}, we can define the one-step-ahead predictive risk as
\[
\bar{L}_T^*(\theta) = \frac{1}{T} \sum_{t=1}^T \mathbb{E}_{t-1}(Y_t - \theta)^2 = \frac{1}{T} \sum_{t=1}^T \mathbb{E}_{t-1}(X_0 + X_t - \theta)^2 = (X_0 - \theta)^2 + 1.
\]
This conditional risk is minimized at \( \theta_T^* = X_0 \). Since the extremum estimator is given by \( \hat{\theta}_T = \frac{1}{T} \sum_{t=1}^T Y_t = X_0 + \frac{1}{T} \sum_{t=1}^T X_t \), we also have that \( |\hat{\theta}_T - \theta_T^*| \to 0 \) almost surely as \( T \to \infty \).

This illustrates that the minimizer of the empirical prequential loss is a consistent estimator of \( \theta_T^* \), which minimizes the conditional predictive risk, instead of the expected risk. When the true data-generating process does not lie within the assumed model class, it is not meaningful to speak of a `true parameter' that generated the observations. In such settings, where models are viewed as predictive systems rather than data-generating mechanisms, consistency should instead be defined in terms of the ability to select a predictor that minimizes the conditional predictive risk.  This is the interpretation of this alternative consistency, first proposed in \cite{skouras2000consistency} for misspecified models.

Moreover, this predictive notion of consistency extends naturally to non-stationary and non-ergodic settings, as shown in Example 4 of \cite{skouras1998optimal}. The authors also emphasize that the choice of loss function plays a critical role: different loss functions may yield different approximations to the true data-generating process (see Example 5 in \cite{skouras1998optimal}). Thus, the choice of the loss function should be guided by the goals of the specific inferential task at hand.

Next, we formally present a ULLN for matingales proved in \cite{skouras2000consistency} under the following conditions. 


\begin{enumerate}[label=\textbf{A\arabic*}]
    \setcounter{enumi}{0}
    \item \label{ChapPPassumption1} The metric space $(\Theta, d)$ is compact.

    \item \label{ChapPPassumption2}  With ${P}$ probability $1$, there exists an $A_t < \infty$ for all $t=1, 2, \cdots$ such that one of the following conditions hold,
    \begin{enumerate}
        \item $$
    \sum_{t=1}^{\infty} \frac{\mathbb{E}_{t-1}\left\{\sup _{\theta \in \Theta}\left|\ell_t(\theta)\right|^2\right\}}{A_t^2}<\infty; \text{ or,}
    $$

    \item there exists an $\epsilon >0$ such that
    $$
    \left(\sum_{t=1}^T \mathbb{E}_{t-1} \{\sup_{\theta \in \Theta} |\ell_t(\theta)|^2\}\right)^{(1+ \epsilon)/2} = \mathcal{O}(A_T).
    $$
    \end{enumerate}

    \item \label{ChapPPassumption3} For each $\theta \in \Theta$, there is a constant $\tau>0$, such that for every $s, d(\theta, s) \leq \tau$ implies that for every $t \geq 1$, ${P}$-a.s.,
    $$
    \left|\ell_t(\theta)-\ell_t(s)\right| \leq B_t\left(Y_t\right) h\{d(\theta, s)\}
    $$
    where $\left\{B_t(\cdot)\right\}$ is a sequence of $\mathcal{F}_{t}$-measurable functions such that ${P}$-a.s.,
    $$
    \limsup _T \frac{1}{A_T} \sum_{t=1}^T \mathbb{E}_{t-1}\left\{B_t\left(Y_t\right)\right\}<\infty
    $$
    and $h(\cdot)$ is a non-random function such that $h(y) \downarrow h(0)=0$ as $y \downarrow 0$. The null sets, $B_t(\cdot)$ and $h$ may depend on $\theta$.
    
\end{enumerate}

\begin{remark}
It is always possible to construct a normalization sequence \(A_T\) that satisfies Assumption~\ref{ChapPPassumption2}. In particular, a natural choice is \(A_t = t\). For a stationary autoregressive processes, we later show that Assumptions~\ref{ChapPPassumption2} and~\ref{ChapPPassumption3} hold under this choice. More generally, this formulation allows for greater flexibility by accommodating data-dependent normalizations, provided that \(A_T\) satisfies Assumptions~\ref{ChapPPassumption2} and~\ref{ChapPPassumption3}.
\end{remark}

\begin{remark}
Clearly, these assumptions are stronger than those typically imposed in the i.i.d.\ setting, where it suffices to assume that the loss is dominated by an integrable function. However, in the present dependent setting, they are useful for controlling the stochastic behaviour and ensuring continuity of the prequential loss.
\end{remark}

\begin{remark}
    As discussed earlier, \cite{pacchiardi2024probabilistic} established consistency of the empirical prequential risk minimizer under asymptotic stationarity, certain mixing conditions that guarantee vanishing dependence between distant observations, and moment boundedness of the loss function. In that sense, assumption~\ref{ChapPPassumption2} can be viewed as analogous to the moment boundedness condition in \cite{pacchiardi2024probabilistic}. This regularity allows our results to extend to settings with stronger temporal dependencies, without requiring either asymptotic stationarity or any form of mixing assumptions.
\end{remark}


With a normalizing constant $A_T$ that satisfies the assumption \ref{ChapPPassumption2}, next we define the normalized prequential loss as, 
$$
\bar{L}_T(\theta) = \frac{1}{A_T} \sum_{t=1}^{T} \ell_t(\theta),
$$
and the normalized expected prequential loss as, 
$$
\bar{L}_T^*(\theta) = \frac{1}{A_T} \sum_{t=1}^{T} \mathbb{E}_{t-1}[\ell_t(\theta)]. 
$$
Note that $A_T$ is taken as $T$ in the example previously discussed from \cite{skouras2000consistency}. Next, we state the ULLN for martingales.

\begin{lemma}\label{ChapPPlem-ullnm}
Under assumptions \ref{ChapPPassumption1}-\ref{ChapPPassumption3}, a uniform law of large numbers (ULLN) for martingales holds, which implies, with probability one under $P$,
$$
\sup_{\theta \in \Theta} \mid \bar{L}_T(\theta) - \bar{L}_T^*(\theta)\mid \rightarrow 0. 
$$
\end{lemma}

Note that \citet{skouras2000consistency} originally proved the above lemma under a different set of assumptions. However, they later introduced assumptions~\ref{ChapPPassumption2} and \ref{ChapPPassumption3} as alternative conditions that are easier to verify in practice. For completeness, we include the original assumptions along with the full lemma statement in Appendix~\ref{ChapPPappendix-lemma1}. Next, we assume that the expected prequential score converges to a finite limit.

\begin{enumerate}[label=\textbf{A\arabic*}]
    \setcounter{enumi}{3}
    \item \label{ChapPPassumption4} There exists a function \( \bar{L}^*(\theta) \) such that as \( T \to \infty \), \( \bar{L}_T^*(\theta) \to \bar{L}^*(\theta) \) uniformly with probability one under $P$.

\end{enumerate}

\begin{corollary}\label{ChapPPcor}
    Under the assumptions \ref{ChapPPassumption1}-\ref{ChapPPassumption4}, it follows from Lemma \ref{ChapPPlem-ullnm} that,
    $$
    \sup_{\theta \in \Theta} |\bar{L}_T(\theta) - \bar{L}^*(\theta)| \to 0 \text{ as } T \to \infty,
    $$
    with probability one under $P$.
\end{corollary}

The above lemma is a direct consequence of Lemma \ref{ChapPPlem-ullnm} and assumption \ref{ChapPPassumption4}. A short proof can be found in Appendix \ref{ChapPPappendix-cor}. We will use this result in addition to the following assumptions to establish a Bernstein-von Mises type theorem on the asymptotic behaviour of the generalized posterior.

\begin{enumerate}[label=\textbf{A\arabic*}]
    \setcounter{enumi}{4}
    \item \label{ChapPPassumption5} The metric space $(\Theta, d)$ is separable.
    \item \label{ChapPPassumption6} For every $t$ and each $\theta \in \Theta$, the functions $\ell_t(\theta)$ is  $\mathcal{F}_t$ measurable, and continuous on $\Theta$ almost surely, \textit{i.e.} it is continuous for all $\omega$ in an event $F_t \in \mathcal{F}_t$ such that $P(F_t) = 1$.

\end{enumerate}

\begin{remark}
    When the assumptions \ref{ChapPPassumption1}, \ref{ChapPPassumption5} and \ref{ChapPPassumption6} hold, then it follows from the following general lemma \citep{gallant1988unified, white1996estimation} that $\bar{L}_T(\theta)$ admits a measurable minimizer. We denote this minimizer by,
$$
\hat{\theta}_T = \arg \min_{\theta \in \Theta} \bar{L}_T(\theta).
$$
We provide the complete statement of the lemma on the existence of the extremum estimator in Appendix \ref{ChapPPlem-existence}.
\end{remark}

 Next, we establish the consistency of the sequence of estimators given by $\{\hat{\theta}_T\}_{T=1}^{\infty}$ to $\theta^*$ defined below.

\begin{enumerate}[label=\textbf{A\arabic*}]
    \setcounter{enumi}{6} 

    \item \label{ChapPPassumption7} The function $\bar{L}^*(\theta)$ has a minimum, $P$-a.s. on $\Theta$ at $\theta^*$. Let $\epsilon > 0$ and $B^c(\epsilon) = \{\theta \in \Theta : d(\theta, \theta^*) \geq \epsilon\}$. Then $P$-a.s.
    $$
 \min_{\theta \in B_T^c(\epsilon)} \bar{L}^*(\theta) - \bar{L}^*(\theta^*) >0.
    $$

\end{enumerate}

\begin{lemma}\label{ChapPPlem-consistency}
    Suppose that assumptions \ref{ChapPPassumption1}-\ref{ChapPPassumption7} hold, then
    $$d(\hat{\theta}_T, \theta^*) \rightarrow 0, $$ 
    with probability one under $P$.
\end{lemma}

We derive the above lemma using a result from \cite{skouras1998optimal} and a detailed proof can be found in Appendix \ref{ChapPPappendix-lem-consistent_estimators}. 

\begin{remark}
    For Lemma \ref{ChapPPlem-consistency} to hold, the parameter space needs to be compact and the empirical prequential score must be smooth enough in the neighbourhood of $\theta^*$ as in condition \ref{ChapPPassumption5}. Then, this result is analogous to the consistency of maximum likelihood estimators (MLE). Specifically, when the loss function is the negative loglikelihood, the estimator \( \hat{\theta}_T \) is the MLE. The result establishes that \( \hat{\theta}_T \) is consistent for \( \theta^* \), the true parameter under a well-specified model, i.e., when \( P = Q^{\theta^*} \). Moreover, as discussed in~\cite{Dawid_2014}, it is possible to construct a statistical divergence induced by the expected prequential loss. Minimizing the expected prequential loss is then equivalent to minimizing this divergence between the true data-generating process and the proposed forecasting model. In the misspecified case, where \( P \notin \mathcal{Q} \), the parameter \( \theta^* \) corresponds to the model within \( \mathcal{Q} \) that minimizes the expected prequential loss, i.e., the model with optimal one-step-ahead predictive performance.
\end{remark}
Using these consistent estimators of \( \theta^* \), we next state the BvM theorem for the generalized prequential posterior under the following assumptions.

\begin{enumerate}[label=\textbf{A\arabic*}]
    \setcounter{enumi}{7}
\item \label{ChapPPassumption8} $\pi : \mathbb{R}^p \rightarrow \mathbb{R}$ is a probability density function with respect to the Lebesgue measure such that $\pi$ is continuous at $\theta^*$ and $\pi(\theta^*) > 0 $.
\item  \label{ChapPPassumption9} $\bar{L}_T(\theta)$ have continuous third derivatives in $E$ and the third derivatives $\bar{L}^{'''}_T(\theta)$ are uniformly bounded in $E$.

    \item \label{ChapPPassumption10} $\bar{L}_T''(\theta^*) \to H^*$  as $T \to \infty$ for some positive definite $H^*$.

    \item \label{ChapPPassumption11} $\theta^*, \hat{\theta}_T \in E$ for sufficiently large $T$, where $E \subseteq \mathbb{R}^p$ is open and convex.

     \item \label{ChapPPassumption12}
     
    For any $\epsilon >0 $, $\lim \inf_T \inf_{\theta \in B^c_{\epsilon}(\hat{\theta}_T)}(\bar{L}_T(\theta) - \bar{L}_T(\hat{\theta}_T)) >0.$

\end{enumerate}

\begin{theorem}\label{ChapPPth-bvm}
    Suppose the assumptions \ref{ChapPPassumption1}-\ref{ChapPPassumption12} hold. Then, there exists a sequence of estimators $\{\hat{\theta}_T\}_{T=1}^{\infty}$ such that as $T \to \infty$, with probability one under $P$, $\hat{\theta}_T \to \theta^*$.
    
    On defining $z_T = \int_{\Theta} \exp(-L_T(\theta)) \pi(\theta)d\theta$ and $\pi_{P}(\theta|y^{T}) = \pi(\theta) \exp(- L_T(\theta))/ z_T$ \footnote{From this point onward, we set $\gamma=1$
    in equation~(\ref{ChapPPeqn: preq-posterior}) wherever the generalized posterior is defined. However, other values of $\gamma$ can be absorbed into the loss function without loss of generality.} we have,
    \begin{equation}
        \int_{B_{\epsilon}(\theta^*)} \pi_{P}(\theta|y^{T}) d\theta \underset{T \rightarrow \infty}{\longrightarrow} 1 \quad \text{for all } \epsilon >0,
    \end{equation}
    which means, $\pi_{P}(\theta|y^{T})$ concentrates at $\theta^*$;
    \begin{equation}
        z_T \approx \frac{\exp(-L_T(\hat{\theta}_T)) \pi(\theta^*)}{\mid \det H^*\mid^{1/2}} \left(\frac{2\pi}{A_T}\right)^{p/2}
    \end{equation}
    as $T \to \infty$ (Laplace approximation), and letting $q_T$ be the density of $\sqrt{A_T}(\theta - \hat{\theta}_T)$ when $\theta \sim \pi_{P}(\theta|y^{T})$,
    \begin{equation}
        \int_{\Theta}\left | q_T(\theta) - \mathcal{N}(\theta \mid \mathbf{0}, {H^*}^{-1}) \right| d\theta \underset{T \rightarrow \infty}{\longrightarrow} 0,
    \end{equation}
    which implies, $q_T$ converges to $\mathcal{N}(\mathbf{0}, {H^*}^{-1})$ in total variation.
\end{theorem}


We prove the above BvM theorem using two results from \cite{JMLR:v22:20-469}, and we provide a sketch of the proof in Appendix \ref{ChapPPappendix-bvm}. The theorem states that as more data is observed, the prequential posterior concentrates around the $\theta^*$. It also begins to resemble a Gaussian distribution. The covariance matrix of this Gaussian is given by the inverse of $H^*$, which plays the role of the Fisher information matrix in the likelihood-based framework.

\subsection{Prequential scoring rule posterior}
\label{ChapPPsubsec: sr}

Previously, we discussed how a discrepancy measure between forecasts from a simulator model and actual observations can define a generalized posterior in the absence of a tractable likelihood function. In this section, we consider scoring rules as a principled way to define such loss functions.

\paragraph{Scoring rules} The scoring rules (SR) originate from the framework of probabilistic forecasting \citep{winkler1967quantification, murphy1970scoring, savage1971elicitation, matheson1976scoring}, where the objective is to quantify the uncertainty associated with a forecast. They are widely used to assess the quality and calibration of probabilistic predictions.
 An SR \( S(Q, y) \) assigns a penalty to a probabilistic model \( Q \) when \( y \) is observed as a realization of the random variable \( Y \). Intuitively, if the model \( Q \) is close to the true underlying process, it incurs a smaller penalty, and vice versa.

Let \( P \) denote the true data-generating process, i.e., \( Y \sim P \). Then, the {expected score} under \( P \) is defined as:
\[
\tilde{S}(Q, P) = \mathbb{E}_{Y \sim P}[S(Q, Y)].
\]
A scoring rule is said to be \textit{proper} if this expected score is minimized when \( Q = P \), and \textit{strictly proper} if the minimum is uniquely achieved at \( Q = P \).

This definition of the properness of scoring rules can be interpreted using the idea of a statistical divergence. Let us define $D(Q, P) = \tilde{S}(Q, P) - \tilde{S}(P, P)$. $S$ is said to be proper if $D(Q, P)$ is a statistical divergence between two probability distributions $Q$ and $P$. This means that $D(P, Q) \geq 0$ for any probability distributions $P$, $Q$. Further, $P = Q$ implies $D(Q, P) = 0$ but the converse only holds when $S$ is strictly proper. This implies that if the true model $P$ is well specified by some model class $\mathcal{Q}$ and the associated scoring rule is proper, then by minimizing the expected scoring rule, one can recover the true model by some $Q^* \in \mathcal{Q}$ where $ Q^* = \arg \min_{Q \in \mathcal{Q}} D(Q, P) = \arg \min_{Q \in \mathcal{Q}} \tilde{S}(Q, P)$. Further, $P$ is the only solution of $Q^*$, if the scoring rule is strictly proper.

Some examples of scoring rules are the continuous ranked probability score (CRPS) \citep{szekely2005new}, energy score or kernel score \citep{gneiting2007strictly} etc. For later sections, we will be using the energy score, which is a multivariate generalization of CRPS, defined as, 
    \begin{equation}\label{ChapPPeqn: esr}
    S_E^{(\beta)}(Q, \mathbf{y}) =2 \cdot \mathbb{E} || X - \mathbf{y}||_2^\beta - \mathbb{E} ||X- X'||_2^\beta, \quad X,  X' \sim Q;\quad \beta \in (0,2).
    \end{equation}
This is a strictly proper scoring rule for the class of probability measures $\mathcal{Q} = \{Q: \mathbb{E}_{X \sim Q} ||X||^{\beta} < \infty, \forall \beta \in (0,2)\}$ \citep{gneiting2007strictly}. The related divergence is the square of the energy distance, which is a metric between probability distributions. Further the energy score can be unbiasedly estimated using $\mathbf{x}_j \sim Q, j=1, \ldots, m$ which are independent and identically distributed samples from the model $Q$, and the estimate is given by,
    \begin{equation}\label{ChapPPeqn: mc-esr}
    \hat{S}_{\mathrm{E}}^{(\beta)}\left(\left\{x_j\right\}_{j=1}^m, \mathbf{y}\right)=\frac{2}{m} \sum_{j=1}^m\left\|x_j-\mathbf{y}\right\|_2^\beta-\frac{1}{m(m-1)} \sum_{\substack{j, k=1 \\ k \neq j}}^m\left\|x_j-x_k\right\|_2^\beta \quad,  \beta \in (0,2).
    \end{equation}
For our implementation, we will consider $\beta = 1$ and will write $S_{\mathrm{E}}^{(1)}(P, \mathbf{y})$ simply as $S_{\mathrm{E}}(P, \mathbf{y})$.

\paragraph{Prequntial scoring rule} The scoring rule-based framework can be extended to handle temporal data as done by \cite{pacchiardi2024probabilistic}. Suppose our simulator model at time \( t \), denoted by \( Q_t^{\theta} \), approximates the conditional distribution of \( Y_t \) given its past values, \( Y_{1:t-1} \). Then, using a scoring rule \( S \), we define the loss as \(\ell_t(\theta) = S(Q_t^{\theta}, Y_t).\)

Upon observing the time series \( Y^T = (Y_1, Y_2, \ldots, Y_T) \), the prequential score is given by, 
\begin{equation}\label{ChapPPeqn: preq-sr}
    L_T(\theta) = \sum_{t=1}^T S(Q_t^{\theta}, Y_t). 
\end{equation}

$L_T(\theta)$ evaluates the performance of the sequence of one step ahead predictions produced by the model $Q^{\theta}$ given the observed history. Consequently, from equation~(\ref{ChapPPeqn: preq-posterior}), we define the generalized prequential scoring rule posterior as,
\begin{equation}
    \pi_{PSR}(\theta \mid y^T) \propto  \pi(\theta) \exp(- L_T(\theta)) = \pi(\theta) \exp\left( - \sum_{t=1}^T S(Q_t^{\theta}, y_t) \right).
\end{equation}


\section{Sequential inference of prequential posteriors}\label{ChapPPsec: sampling scheme}
 In this section, we describe our sampling strategy for a high-dimensional parameter space (with dimension $\sim 1000$) to sample from the prequential posterior distribution defined previously. First, we define an episode of length $\tau$ as $Y_{\tau(i-1) + 1: \tau i}$, where $i = 1, 2, \ldots$. These episodes can be viewed as an update schedule, since it is often undesirable to update model parameters upon the arrival of every single datapoint. Instead, we choose an update schedule based on a fixed episode length, after which we recompute the posterior distribution. With a slight abuse of notation, we write the posterior distribution of the model parameters after observing $Y^{\tau t}$ as
\[
\pi_t(\cdot) = \pi_{PSR}(\cdot \mid y^{\tau t}).
\]
This defines a sequence of target distributions from which we want to sample. The sequence is given by $\{\pi_1(\cdot), \pi_2(\cdot), \ldots\}$, corresponding to the posterior distributions after observing the data series $Y^\tau, Y^{2\tau}, \ldots$. To sample from this sequence, we propose the use of Sequential Monte Carlo (SMC) samplers \citep{del2006sequential}.

\subsection{Sequential Monte Carlo}
SMC provides a powerful class of sampling algorithms that originated from particle filters used in state-space models. These samplers have since been extended to a wide range of statistical problems involving complex target distributions. The key idea is to use a set of particles and propagate them through the sequence of target distributions starting from a proposal distribution. Traditional sampling methods, such as importance sampling and Markov chain Monte Carlo (MCMC), are well-established but may face limitations in high-dimensional or dynamic settings. SMC samplers integrate the strengths of both these methods in a way that enhances its scalability.

In practice, SMC approximates the sequence of target distributions using a population of particles that evolve through a combination of MCMC moves, importance sampling, and resampling to avoid degeneracy. Let us consider the Markov kernel $M_t(\theta_{t-1}, d\theta_t)$ such that it is $\pi_{t-1}$ invariant which means, $\pi_{t-1} M_t = \pi_{t-1}$ for $t \geq1$, where we use the notation $\theta_t$ to denote a particle at $t$-th iteration of SMC. We can then define the following sequence of Feynman-Kac distributions for $t=0, 1, 2, \ldots$ as,

\begin{equation}
    \mathbb{F}_t(d\theta_{0:t}) = \frac{1}{z_t} \pi(d\theta_0)\prod_{s=1}^t M_s(\theta_{s-1}, d\theta_s) \prod_{s=0}^t G_s(\theta_s).
\end{equation}
Here, the potential function is given by 

$$G_s(\theta) = \exp (L_{s\tau}(\theta)/L_{s(\tau - 1)}(\theta)) = \exp(\sum_{t = s(\tau - 1) +1}^{s\tau} l_t(\theta)).$$ 
The construction above ensures that the marginal distribution of $\theta_t$ (with respect to $\mathbb{F}_t$) is $\pi_t$. We use this formulation to set up an SMC sampler with MCMC moves, following the approach suggested in \cite{del2006sequential}[Section 3.3.2.3]. In practice, however, it is often challenging to employ an exact MCMC kernel with the correct invariant distribution, since the Metropolis correction step is frequently omitted to reduce computational cost and to ensure mixing. The details of our Markov kernel implemented are provided in Section \ref{ChapPPsec: precondition}.




\subsection{Adaptive tempering within SMC}\label{ChapPPsec-kernel}
It is important for the efficiency of SMC samplers that the proposal and target distributions are sufficiently close to avoid weight degeneracy. To address this, we utilize an adaptive strategy based on the \emph{conditional effective sample size} (CESS) \citep{zhou2016toward}, which automatically determines the sequence of intermediate distributions to ensure smooth propagation of particles within each episodic update. Intuitively, CESS quantifies the overlap between two successive distributions, which can be evaluated using the incremental weights corresponding to a hypothetical transition between these distributions.

In our framework, since the incremental weight can be computed prior to the forward propagation of particles, we enforce a constant overlap across successive intermediate distributions by keeping the CESS fixed. This avoids degeneracy during weight computation. To transition particles from $\pi_{t-1}(\cdot)$ to $\pi_t(\cdot)$, we define a sequence of intermediate distributions as:
\[
\left\{ \pi_t^{(l)}(\theta) = \pi_{t-1}(\theta)^{1 - \lambda_l} \pi_t(\theta)^{\lambda_l} \right\}_{l=1}^L,
\]
where the \emph{temperatures} $\{\lambda_l\}$ are such that $0 < \lambda_1 < \lambda_2 < \cdots < \lambda_L = 1$. These temperature values are adaptively chosen so that the CESS remains constant across transitions, thereby ensuring a smooth evolution of the particle system through the episodic posterior updates.

\subsection{Wastefree SMC}
We previously discussed how SMC samplers employ forward kernels to propagate particles from a proposal distribution towards the target distribution. In high-dimensional settings, it is critical that these kernels exhibit strong mixing properties to ensure adequate exploration of the sample space. In practice, when sampling from a high-dimensional parameter space, it is common to apply multiple iterations of MCMC moves to reduce the variance. However, only the final sample in the chain is typically retained as the next particle, while the intermediate samples are discarded. This seems like a loss of potentially valuable computational effort.

\cite{dau2022waste} proposed a strategy to mitigate this inefficiency by utilizing \emph{all} intermediate samples generated by the MCMC chains. This approach enables a larger effective number of particles to represent the target distribution, without incurring a prohibitive increase in computational cost. 

The key idea is as follows: suppose that we aim to generate $N$ particles to represent the posterior distribution at each update step. Instead of sampling all $N$ particles independently, we first resample $M \ll N$ particles from the current population. From each of these $M$ particles, we initiate an MCMC chain of length $P$, resulting in a total of $N = MP$ particles when all chain samples are collected. This strategy is particularly advantageous in parallel computing environments where each of the $M$ MCMC chains can be executed simultaneously on $M$ cores. This offers a computationally efficient means of generating a diverse particle population.

Furthermore, the authors provide a Feynman–Kac model interpretation of the \emph{waste-free} SMC algorithm. The fundamental difference from standard SMC lies in the structure of the target space: rather than sampling in $\Theta$, the algorithm samples paths in the extended space $\Theta^P$, with each path corresponding to a full MCMC trajectory of length $P$. They exploit this duality to establish theoretical guarantees, including the consistency and asymptotic normality of the particle approximation.

\subsection{Preconditioned Forward Kernel}\label{ChapPPsec: precondition}

Efficient sampling becomes increasingly challenging as the dimensionality of the parameter space grows, a phenomenon commonly referred to as the \emph{curse of dimensionality}. This issue affects most sampling methods, including SMC, MCMC, and their variants. To address this, gradient-based Markov kernels \citep{welling2011bayesian, chen2014stochastic, ding2014bayesian} are often employed to enhance exploration of the parameter space by leveraging gradient information.

One widely used gradient-based approach is \emph{stochastic gradient Langevin dynamics} (SGLD)~\citep{welling2011bayesian}, which generates samples by discretizing a stochastic differential equation (SDE). The Euler–Maruyama scheme is typically used for discretization due to its simplicity, although it introduces discretization bias. Further, to mitigate the effect of noisy gradient estimates, momentum and friction terms can be incorporated, leading to the \emph{stochastic gradient Nosé–Hoover thermostat} (SGNHT) algorithm~\citep{leimkuhler2016adaptive}. This method operates in an extended state space that includes both parameters and auxiliary momentum variables, and has found successful application in Bayesian inference~\citep{ding2014bayesian, pacchiardi2024generalized, roy2024generalized}.

Additionally, preconditioning strategies have been proposed to further improve sampling efficiency in high-dimensional spaces. For example, \citet{girolami2011riemann} introduced Riemannian geometry-based preconditioning, while \citet{chen2016bridging} proposed using the squared gradient of the loss, akin to the preconditioners used in adaptive optimizers like RMSprop and Adam. We adopt the latter approach in our work. Furthermore, to improve numerical accuracy when solving the SDE, we use a symmetric splitting scheme (as shown in \cite{chen2016bridging}), which provides better approximation quality at the cost of slightly increased computation.

The following pseudo-algorithm outlines our forward Markov kernel for sampling from a distribution with density proportional to \( \pi(\theta) \cdot \exp\{-L_T(\theta)\} \), where \( L_T(\theta) \) denotes the (prequential) loss and \( \pi(\theta) \) is the prior. The corresponding potential function is defined as 
\[
U(\theta) = L_T(\theta) - \log \pi(\theta).
\]
Our method is inspired by the Santa-SSS algorithm introduced in~\cite{chen2016bridging}. We provide more details on the kernel in Appendix~\ref{ChapPPappendix:santa}.

\begin{algorithm}[H]
\caption{Preconditioned forward kernel}
\label{ChapPPalg:santa-sss}

\KwIn{Learning rate $\eta$, parameters $(\sigma_1, \sigma_2)$, noise $\{ \boldsymbol{\zeta}_t \sim \mathcal{N}(\mathbf{0}, \mathbf{I}_p) \}$}
\KwInit{$\boldsymbol{\theta}_0$, $\mathbf{u}_0 = \sqrt{\eta} \cdot \mathcal{N}(0, \mathbf{I}_p)$, $\boldsymbol{\alpha}_0 = \sqrt{\eta} \cdot C$, $\mathbf{v}_0 = \mathbf{0}$}

\For{$n = 1, 2, \ldots $}{
    Compute stochastic gradient: $\tilde{\mathbf{f}}_n \gets \nabla_{\boldsymbol{\theta}} \tilde{U}(\boldsymbol{\theta}_{n-1})$\;

    $\mathbf{v}_n \gets \sigma \mathbf{v}_{n-1} + \frac{1 - \sigma}{T^2} \tilde{\mathbf{f}}_n \odot \tilde{\mathbf{f}}_n$\;

    $\mathbf{g}_n \gets 1 \oslash \sqrt{\sigma_2 + \sqrt{\mathbf{v}_n}}$\;

    $\boldsymbol{\theta}_n \gets \boldsymbol{\theta}_{n-1} + \frac{1}{2} \mathbf{g}_n \odot \mathbf{u}_{n-1}$\;

    $\boldsymbol{\alpha}_n \gets \boldsymbol{\alpha}_{n-1} + \frac{1}{2} \left( \mathbf{u}_{n-1} \odot \mathbf{u}_{n-1} - \eta \right)$\;

    $\mathbf{u}_n \gets \exp(-\boldsymbol{\alpha}_n / 2) \odot \mathbf{u}_{n-1}$\;

    $\mathbf{u}_n \gets \mathbf{u}_n - \eta \cdot \mathbf{g}_n \odot \tilde{\mathbf{f}}_n + \sqrt{2 \cdot \mathbf{g}_{n-1} \eta^{3/2}} \odot \boldsymbol{\zeta}_n$;

    $\mathbf{u}_n \gets \exp(-\boldsymbol{\alpha}_n / 2) \odot \mathbf{u}_n$\;

    $\boldsymbol{\alpha}_n \gets \boldsymbol{\alpha}_n + \frac{1}{2} \left( \mathbf{u}_n \odot \mathbf{u}_n - \eta \right)$\;

    $\boldsymbol{\theta}_n \gets \boldsymbol{\theta}_n + \frac{1}{2} \mathbf{g}_n \odot \mathbf{u}_n$\;
}
\textbf{end}
\end{algorithm}
We employ the above Markov kernel in Algorithm~\ref{ChapPPalg:santa-sss} within the waste-free SMC framework. Note that we do not use a Metropolis accept--reject step; instead, the next particle is always accepted. While this introduces an approximation error, it reduces computational overhead.  

Moreover, to obtain an unbiased estimate of the prequential loss $L_T(\theta)$, consider an alternative representation of the energy score:  
\[
\ell_t(\theta) = S(Q^{\theta}_t, Y_t) = \mathbb{E}_{X, X' \sim Q^{\theta}_t} \, g(X, X', Y_t),
\]  
for some function $g$. $Q^{\theta}_t$ being a DGFM, a simulation from the model can be expressed as $X = h_{\theta}(W_1)$, where the function $h_{\theta}$ is defined by deep neural networks parameterized by $\theta$ and $W_1$ is drawn from an auxiliary distribution $W$, independent of $\theta$, as shown in ~\cite{pacchiardi2024generalized}. Under mild regularity conditions, and assuming both $h_{\theta}$ and $g$ are differentiable with respect to $\theta$, we can interchange expectation and differentiation, yielding:  
\begin{align*}
\nabla_{\theta} S(Q^{\theta}_t, Y_t) 
&= \nabla_{\theta} \, \mathbb{E}_{X, X' \sim Q^{\theta}_t} g(X, X', Y_t) \\ 
&= \nabla_{\theta} \, \mathbb{E}_{W_1, W_2 \sim W} g(h_{\theta}(W_1), h_{\theta}(W_2), Y_t) \\ 
&= \mathbb{E}_{W_1, W_2 \sim W} \, \nabla_{\theta} g(h_{\theta}(W_1), h_{\theta}(W_2), Y_t),
\end{align*}
where the gradient can be calculated using automatic-differentiation. This explains how we unbiasedly estimate the score using independent draws from the base distribution $W$. 


\subsection{Complete sequential inference procedure}

\begin{algorithm}[htbp!]
\caption{Sequential inference for prequential posteriors via wastefree SMC}
\label{ChapPPalg:prequential_wf_smc}

\DontPrintSemicolon
\SetKwInOut{Input}{Input}
\SetKwInOut{Output}{Output}

\Input{
Observed sequence $y_{1:T}$ partitioned into episodes of length $\tau$; \\
the loss function $\ell(\cdot)$ (or, an unbiased estimate $\hat{\ell}(\cdot)$, if $\ell(\cdot)$ is intractable;\\
Prior $\pi_0(\theta)$; \\
Number of retained trajectories $M$; \\
Number of kernel iterations $P$ (total particles $N=MP$); \\
CESS threshold $\kappa N$ where $\kappa \in (0, 1)$; \\
Forward kernel $K_{\eta}$ (Algorithm~\ref{ChapPPalg:santa-sss})
}

\Output{
Weighted particle approximation of
$\{\pi_t(\theta)\}_{1:\lceil T/\tau \rceil}$
}

Initialize particles
$\{\theta_0^{(n)}\}_{1:N}\sim\pi_0(\theta)$,
weights $w_0^{(n)}\leftarrow 1/N$\;

\For{$t=1,\dots,\lceil T/\tau\rceil$}{

Define current episodic loss: $\Delta L_t(\theta)
=
\sum_{s=(t-1)\tau+1}^{t\tau}
\ell_s(\theta)$

Define target posterior
\[
\pi_t(\theta)
\propto
\pi_{t-1}(\theta)
\exp\{-\Delta L_t(\theta)\}
\]

Set temperature index $l\leftarrow 0$, $\lambda_0=0$, $\theta^{(n, 0)} = \theta_{t-1}^{(n)}, w^{(n, 0)} = w_{t-1}^{(n)} $\;

\While{$\lambda_l < 1$}{

Choose next temperature
$\lambda_{l+1}\in(\lambda_l,1]$
such that $\mathrm{CESS}(\lambda_l,\lambda_{l+1})=\kappa N$

Define the intermediate target density
$$
\pi_t^{(l)}(\theta) = \pi_{t-1}(\theta) \exp\left(
-(\lambda_{l+1}-\lambda_l)
\Delta L_t(\theta)\right)
$$

Resample $M$ ancestor particles
according to
$ w^{(n, l)}$\;

\For{$m=1,\dots,M$}{

Initialize chain:
\[
\theta_{m,0}
\leftarrow
\texttt{Resample}(1, \{\theta^{(n, l)}\}_{1:N}, \{w^{(n, l)}\}_{1:N})
\]

\For{$p=1,\dots,P$}{

Propagate: $\theta_{m,p}
\sim
K_{\eta}(
\theta_{m,p-1},
\pi_t^{({l+1})}
)$

}

}

Flatten all chains and update the weights:
\[
\{
\theta^{(n, l)}
\}_{1:N}
=
\{
\theta_{m,p}
\}_{1:M, 1:P}
\]

\begin{equation}\label{ChapPPeqn: smc_weight update}
    \{\widetilde w^{(n, l)}\}_{1:N}
\leftarrow
\{\exp\left(
-(\lambda_{l+1}-\lambda_l)
\Delta L_t(\theta_{m, p-1})
\right)\}_{1:M, 1:P}
\end{equation}

Normalize weights: $
w^{n,l} = \widetilde w^{(n, l)}/\sum_{n=1}^N \widetilde w^{(n, l)}
$

$l\leftarrow l+1$\;

}

Gather the particles and the weights
$$
\{\theta_t^{(n)} = \theta^{(n, l+1)}\}_{1:N}, \quad  \{w_t^{(n)} = w^{(n, l+1)}\}_{1:N}
$$


}

\end{algorithm}

Section~\ref{ChapPPsec: gen bayes} establishes consistency and Bernstein--von Mises behaviour for the exact prequential posterior $\pi_P(\theta| y_{1:T}) \propto \pi(\theta)\exp\{-L_T(\theta)\}$. However, exact sampling from this target is infeasible for deep generative models and we target the approximate density $\hat{\pi}_P(\theta| y_{1:T}) \propto \pi(\theta)\exp\{-\hat{L}_T(\theta)\}$ instead, where $\hat{L}_T(\theta)$ is an unbiased estimate of $L_T(\theta)$. However, \cite{pacchiardi2024generalized} showed that with enough simulations to estimate $L_T(\theta)$ the gap between the exact and the approximate posterior vanishes under some regularity conditions. We now consolidate all the components of the proposed SMC sampler 
designed to target this posterior in Algorithm~\ref{ChapPPalg:prequential_wf_smc}.

Note that the weight update equation (Equation~\ref{ChapPPeqn: smc_weight update}) expresses the weight of the $p$-th particle as a function of its previous state ($p-1$). This follows the standard MCMC kernel construction (see Section~3.3.2.3 in~\cite{del2006sequential}). Importantly, this formulation allows the incremental importance weights to be evaluated prior to the mutation step. As a result, the sequence of temperatures $\{\lambda_l\}$ are adaptively determined based on the current population of particles under the CESS constraint before applying the mutation kernel.

\section{Prequential posteriors in practice} \label{ChapPPsec: sim study}

We developed a sequential inference method for the prequential posteriors in Section~\ref{ChapPPsec: sampling scheme}. Next, we demonstrate our proposed data assimilation strategy on three benchmark datasets: a stationary autoregressive process of order $q$ (AR($q$)), a time series model for weather forecasting introduced in \cite{lorenz1996predictability}, and an open-source meteorological benchmarking dataset called \textit{WeatherBench}. The AR($q$) example represents a well-specified setting, as the simulator is also 
an autoregressive process and the likelihood is tractable; we nevertheless use the prequential posterior here for illustration. In the other two examples, we employ a deep generative forecasting model as the simulator. For Lorenz 96, we also compare our method with the ensemble Kalman filter, assuming a commonly used but misspecified state-space model. ABC-SMC-type algorithms would have been a natural benchmark in this setting; however, we do not include them due to well-known scalability limitations of ABC-based sampling methods in high-dimensional problems. As the true likelihood function is intractable for such models, we instead define the prequential posterior using energy score as the loss function. Specifically, we use the energy score (Equation~\ref{ChapPPeqn: esr}), and obtain an unbiased estimate of the prequential score via the Monte Carlo approximation in Equation~\ref{ChapPPeqn: mc-esr}, applied to the prequential form given in Equation~\ref{ChapPPeqn: preq-sr}.

To incorporate new data sequentially, we partition the training data into fixed-length episodes of size \( \tau \); where we have taken $\tau = 10$ for the AR process, $\tau = 100$ for Lorenz 96 and $\tau = 362 (\sim 1$ year) for WeatherBench. The generalized posterior is updated at the end of each episode using an SMC sampler. The proposal distribution is taken to be a multivariate standard Gaussian, though we also examined the effect of alternative choices (e.g., Student’s t-distribution) for the deep generative models, with results presented in the Appendix \ref{ChapPPappendix: prior study}. In all cases, the prior is chosen to be identical to the proposal. The tempering of the episodic posteriors is performed by keeping the conditional effective sample size (CESS) fixed to a threshold, ensuring adaptive yet stable intermediate distributions. Full implementation details are provided in the Appendix \ref{ChapPPappendix:smc table}.

We assess the predictive performance of the episodic SMC samples on a held-out test dataset. For each episodic posterior, we evaluate the quality of the probabilistic forecasts using a probabilistic calibration metric (\textit{calibration error}) and two deterministic metrics: the \textit{normalized root mean squared error} (NRMSE) and the \textit{coefficient of determination} (\( \mathrm{R}^2 \)). A detailed description of these metrics is provided in Appendix~\ref{ChapPPappendix-calerr}.

\subsection{A stationary autoregressive process}\label{ChapPPsec: arq_process}
We consider a time series following a stationary autoregressive process of order 
$q = 5$, so that the true data-generating process is
\begin{align*}
    Y_t &= \theta^{*\prime} Y_{t-q:t-1} + \epsilon_t, \qquad \epsilon_t \sim \mathcal{N}(0,1), \\
    \theta^* &= [\theta_1^*, \dots, \theta_q^*]',
\end{align*}
where $\theta^*  = [0.6689,\ 0.0401,\ {-0.4582},\ 0.9265,\ {-0.1972}]'$. We discuss the derivation of true coefficients in Appendix~\ref{ChapPPappendix: stationary coefficients ar5}.

\paragraph{Autoregressive forecasting model}
We define the model class $\mathcal{Q}$ as the family of AR($q$) processes with 
$q = 5$, so that the true data-generating process is well-specified. The simulator 
is given by
\begin{align*}
    X_t &= \theta' Y_{t-q:t-1} + \omega_t, \qquad \omega_t \sim \mathcal{N}(0,1),\\
    \theta &= [\theta_1, \dots, \theta_q]',
\end{align*}
and the autoregressive coefficient vector $\theta$ is the parameter of interest. 
We define the prequential posterior to perform inference on $\theta$.

\paragraph{Verification of the regularity conditions}
The consistency results of Section~\ref{ChapPPsec: gen bayes} require regularity 
conditions on both the true process $P$ and the loss function $\ell_t(\theta)$. 
We verify the assumptions~\ref{ChapPPassumption2}-\ref{ChapPPassumption4}  which are relevant to the `prequential' framework  for the present example, with detailed proofs deferred 
to the Appendix~\ref{ChapPPappendix-assumption_check}. We show that the conditions hold with 
$A_t = t$. The key step is bounding the relevant conditional expectations, which impose minimal restriction on $P$. These assumptions are verified via 
stationarity and ergodicity of the observed process. We highlight that ergodicity 
is a strictly weaker condition than $\alpha$-mixing or $\beta$-mixing. It does not 
require the temporal correlations to decay, and the assumption therefore remains 
valid even under strong dependence.

\paragraph{Empirical findings}

\begin{figure}
    \centering
    \includegraphics[width=0.95
    \linewidth]{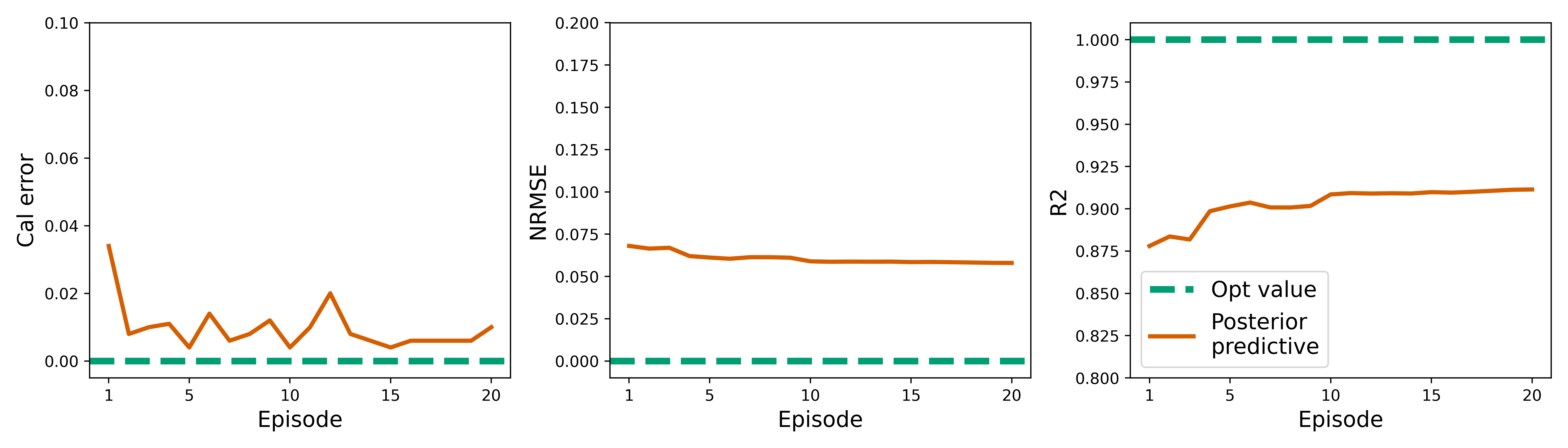}
    \caption{Data assimilation for the stationary AR($5$) process. We infer the prequential posteriors on observing new $\tau= 10$ observations and plot the diagnostics computed on a separate test data of length $300$. The orange curve shows the value of
    the calibration metrics (calibration error, NRMSE, $R^2$) of the posterior predictives obtained after each assimilation episode. The optimum values of the metrics are shown in the green dashed line.}
    \label{ChapPPfig:ar5_episodic_calibration}
\end{figure}

For the AR($q$) process experiment, we infer prequential posteriors after each episode, where an episode consists of $\tau$ observations. We then evaluate the predictive performance of the posterior samples on a test dataset of length $300$ in terms of calibration error, NRMSE, and the coefficient of determination $R^2$. Figure~\ref{ChapPPfig:ar5_episodic_calibration} shows as we gather more episodic data, both the calibration error and the normalised RMSE of the posterior predictive decrease and approach $0$ while $R^2$ increases towards $1$. This behaviour indicates effective data assimilation, with predictive performance improving progressively as the posterior distribution is updated with additional observations.

As discussed in Section~\ref{ChapPPsec: precondition}, we do not employ the Metropolis-Hastings correction step within the Markov kernel. Nevertheless, in Appendix~\ref{ChapPPappendix:sensitivity}, we investigate the effect of retaining the accept-reject correction on the convergence behaviour of the posterior samples via kernel Stein discrepancy (KSD) \citep{gorham2017measuring}, across different learning rates ($\eta$). Our results indicate that the proposed kernel without the Metropolis--Hastings correction is more robust to the choice of learning rate and achieves the best performance for $\eta \in [10^{-3}, 10^{-1}]$. Here we report results corresponding to the learning rate $\eta = 0.01$.

We also examine the sensitivity of posterior convergence to the total number of particles $(N)$ and the Markov chain length $(P)$ used in waste-free SMC; details are provided in Appendix~\ref{ChapPPappendix:sensitivity}. As expected, performance improves with increasing $N$, as a larger particle population reduces estimation variance. We additionally observe improved performance with longer Markov chains, which is consistent with the findings of \citep{dau2022waste}, where the authors recommend choosing $M \ll N = MP$, thereby favouring larger values of $P$.

However, it is important to note that Markov chain propagation is inherently sequential and cannot be parallelized efficiently. However, increasing the number of resampled particles $(M)$ allows parallel execution and may therefore offer computational advantages in practice. Based on our sensitivity analysis, we find that fixing $N = 150$ and choosing $P = 10$ provides satisfactory performance, and the reported result use this configuration. Additional details and complete sensitivity analyses are provided in Appendix~\ref{ChapPPappendix:sensitivity}.

\subsection{Lorenz 96}


The Lorenz96 model~\citep{lorenz1996predictability} is a simplified, yet widely used system designed to mimic key features of atmospheric dynamics. It consists of two types of variables: slow-evolving variables \( y\) and fast-evolving variables \( x\) and we only record the slow-evolving $y$ variable as observations. The coupled evolution of these variables is governed by the following set of nonlinear differential equations:

\begin{align}\label{ChapPPeqn-true-l96}
\dot{y}(k) & = -y{(k-1)} \Bigl( y({k-2}) - y({k+1}) \Bigr) - y(k) + F - \frac{h c}{b} \sum_{j=J(k-1)+1}^{kJ} x(j) \notag \\
\dot{x}(j) & = -c b x({j+1})\Bigl(x({j+2}) - x({j-1})\Bigr) - c x(j) + \frac{h c}{b} Y_{\lfloor (j-1)/J \rfloor + 1}
\end{align}
where $y(k)$ and $x(j)$ denote the $k$th and the $j$th component of the respective variables. Here, \(k = 1, \ldots, K\) and \(j = 1, \ldots, JK\), with cyclic boundary conditions applied such that \(k = K+1\) maps to \(k = 1\), and similarly for \(j\). These equations capture the interactions between slow and fast variables in a structured, recurrent fashion—each \( y \) is influenced by a group of \( J \) fast components \( x \).

Following the experimental setup in~\citet{pacchiardi2024probabilistic}, we fix the model parameters to \( K = 8 \), \( J = 32 \), \( h = 1 \), \( b = 10 \), \( c = 10 \), and \( F = 20 \). We integrate the system using the fourth-order Runge–Kutta (RK4) method with a discretization step of \( \mathrm{d}t = 0.001 \). The initial conditions are set to \( y(1) = x(1) = 1 \), while all other components are initialized to zero. After discarding the first 2 units of simulated time (to allow transients to settle), the values of \( {y} \) are recorded every \( \Delta t = 0.2 \), roughly corresponding to one atmospheric day in terms of predictive dynamics. Simulation is carried out for $4000$ additional time units. The resulting dataset is divided into training and test portions using an $80\%-20\%$ split.

\paragraph{The deep generative forecasting model (DGFM)}

We define our model class \( \mathcal{Q} \) as a family of generative models with a Markov dependency structure of order \(  10 \). This implies that the prediction at time \( t \) depends on the preceding \( 10 \) observations.

The generative model architecture takes an input sequence \( Y_{t-k:t-1} \), which is fed into a single-layer gated recurrent unit (GRU) network with a hidden size of 16. The output from the GRU is concatenated with a latent noise vector \( W \sim \mathcal{N}(0,1) \), where \( W \in \mathbb{R}^1 \). This combined feature representation is then passed through a stack of three fully connected (dense) layers, which outputs the prediction for the next time step \( Y_t \), for \( t = k+1, k+2, \ldots \). The total number of trainable parameters in this simulator amounts to \( p = 4442 \).

\paragraph{Ensemble Kalman filter with a misspecified state-space model}
To benchmark our method against a standard baseline, we employ the ensemble Kalman filter (EnKF). 
The following state-space model is commonly used for inference in the Lorenz--96 system 
(see, for example, \cite{Roth_2017} and \cite{duranmartin2024outlierrobustkalmanfilteringgeneralised}), 
where \(x\) denotes the latent state and \(y\) the noisy observation:
\begin{align} \label{ChapPPeqn-simple-l96}
    \dot{x}(k) &= \bigl(x(k+1) - x(k-2)\bigr)x(k-1) - x(k) + F_k, \\ \notag
    y(k) &= x(k) + \psi_k,
\end{align}
with \(F_k \sim \mathcal{N}(20, 1)\) and \(\psi_k \sim \mathcal{N}(0, 1)\) for \(k = 1, 2, \ldots, K\), where \(K = 8\). Note that this simplified model is obtained by isolating the slow-variable equation from 
\eqref{ChapPPeqn-true-l96} and treating the forcing term \(F\) as known, following 
\cite{lorenz1996predictability}. Nevertheless, the model remains misspecified relative to 
the full system, since the assumed Gaussian observation noise cannot capture the chaotic 
variability introduced by the fast variables. We then discretize \eqref{ChapPPeqn-simple-l96} 
using a fourth-order Runge--Kutta (RK4) method, yielding the nonlinear state transition 
equations employed by the EnKF.

\begin{figure}[htbp]
    \centering
    \includegraphics[width=0.95\linewidth]{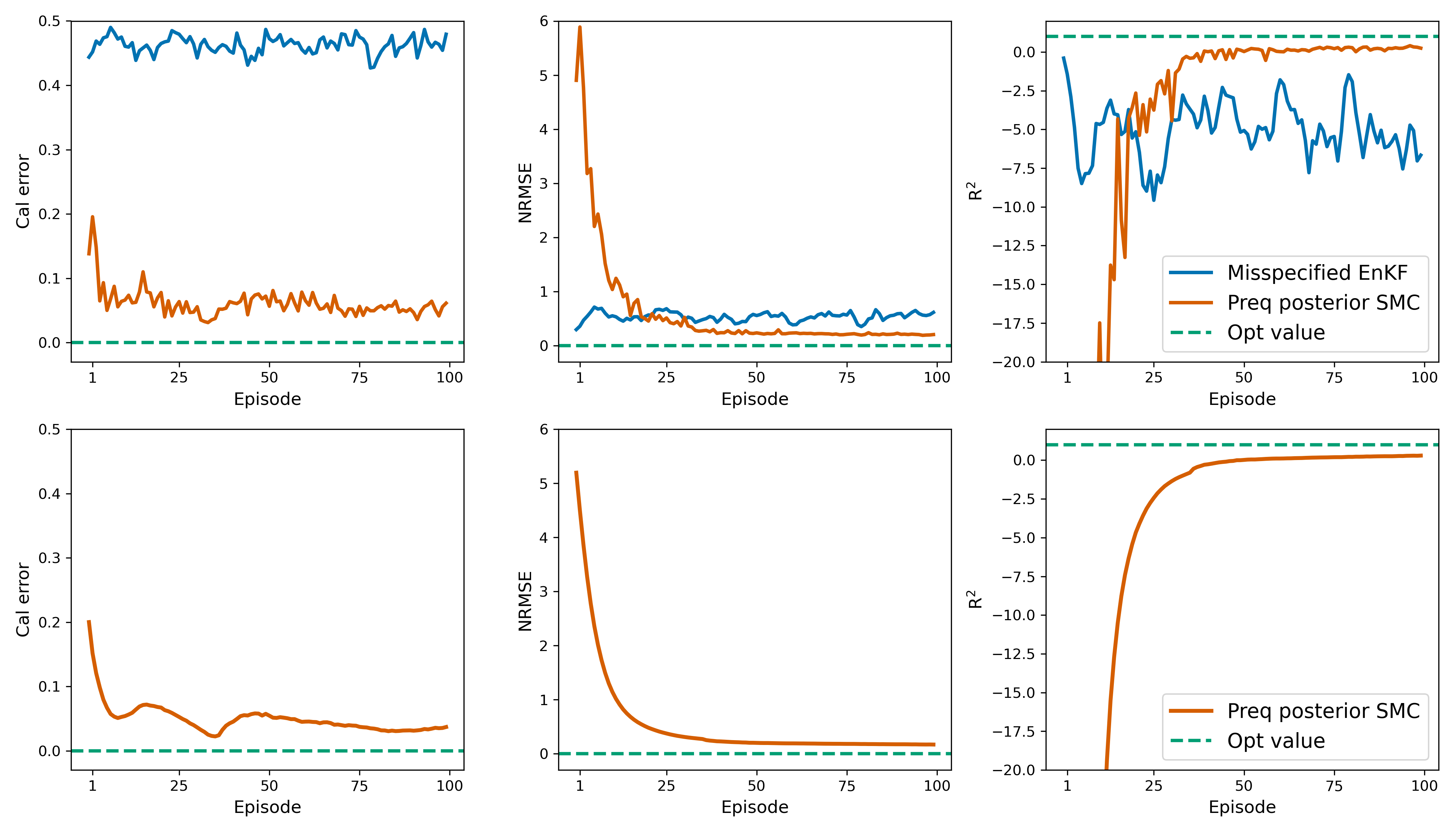}
    \caption{Posterior predictive performance across episodes for the Lorenz 96 model. The prequential posterior (orange curve) improves with additional data, while the misspecified EnKF (blue curve) shows little change, as measured by calibration error, normalised RMSE, and coefficient of determination $R^2$. The green dotted line represents the maximum achievable value for each metric. In the top panel, diagnostics are computed on the next episode (of length $\tau = 100$) of the same dataset, while the bottom panel uses a separate test dataset (of length $2000$).}
    \label{ChapPPfig:calerr}
\end{figure}

\paragraph{Empirical findings}
We infer the prequential posterior after each episode and assess its predictive performance on a separate dataset using calibration error, NRMSE, and \(R^2\). In the top panel of Figure~\ref{ChapPPfig:calerr}, we evaluate the predictive accuracy of the posterior 
obtained after the \(i\)th episode on the data segment \((\tau i + 1,\ (\tau + 1)i)\) 
for \(i = 1, \ldots, 100\). 
For comparison, the EnKF continually updates its predictive distribution at every time step, 
and we compute the same diagnostic metrics. 
Because the state-space model assumed by the EnKF is misspecified, its predictive accuracy does not 
improve even as more data are observed. 
In contrast, our method progressively learns a model that better represents the underlying 
data-generating mechanism, leading to substantial improvement in predictive accuracy: 
although initially poor, it eventually surpasses the performance of the EnKF.

In the bottom panel of Figure~\ref{ChapPPfig:calerr}, the diagnostic metrics are evaluated on a 
separate test set of length \(2000\). 
As additional episodes are incorporated, both the calibration error and the normalized RMSE decrease and approach zero, indicating enhanced calibration and accuracy. Simultaneously, the coefficient of determination \( R^2 \) increases towards $1$, which aligns with the theoretical result according to Theorem~\ref{ChapPPth-bvm} on posterior consistency.




\subsection{WeatherBench}
The WeatherBench dataset \citep{rasp2020weatherbench} is an open-source benchmark for data-driven weather forecasting. It provides hourly measurements of various atmospheric fields from 1979 to 2018 at multiple spatial resolutions. For this example, deterministic forecasting systems such as Pangu-Weather~\citep{bi2022panguweather3dhighresolutionmodel}, FourCastNet~\citep{pathak2022fourcastnetglobaldatadrivenhighresolution}, GraphCast~\citep{lam2023graphcastlearningskillfulmediumrange}, and ClimaX~\citep{pmlr-v202-nguyen23a}, as well as DGFM approaches such as GenCast~\citep{price2024gencastdiffusionbasedensembleforecasting} and~\cite{pacchiardi2024generalized}, can be applied, but they typically rely on point estimates. In this section, we instead provide a Bayesian alternative to these forecasting methods with a natural application to data assimilation.

In this work, we select a resolution of $5.625^{\circ}$ in both longitude and latitude, corresponding to a $32 \times 64$ grid, and focus on an $8 \times 16$ subset covering the European subregion. We use one observation per day (12:00 UTC) of the 500 hPa geopotential height (Z500) field.
Our forecasting task is to predict one day ahead using the previous three daily observations. The period $1979-1998$ is used for training, and calibration tests are conducted on the $1999-2002$ data.

\paragraph{The deep generative forecasting model (DGFM)}
We adopt a U-Net architecture \citep{ronneberger2015u} for the generative network, following the adaptation in \cite{pacchiardi2024probabilistic}. This encoder–decoder structure processes the input through successive encoder layers, each producing a downscaled latent representation. The final encoder output is passed to a bottleneck layer, which preserves the scale while transforming the features. The decoder then progressively upsamples this representation, with skip connections linking encoder and decoder layers at the same resolution. These connections ensure that both large-scale structures and fine-scale details contribute to the output. A random noise is drawn from standard Gaussian distribution and added to the bottleneck representation before decoding.

\paragraph{Empirical findings}

\begin{figure}[htbp]
    \centering
    \includegraphics[width=0.95\linewidth]{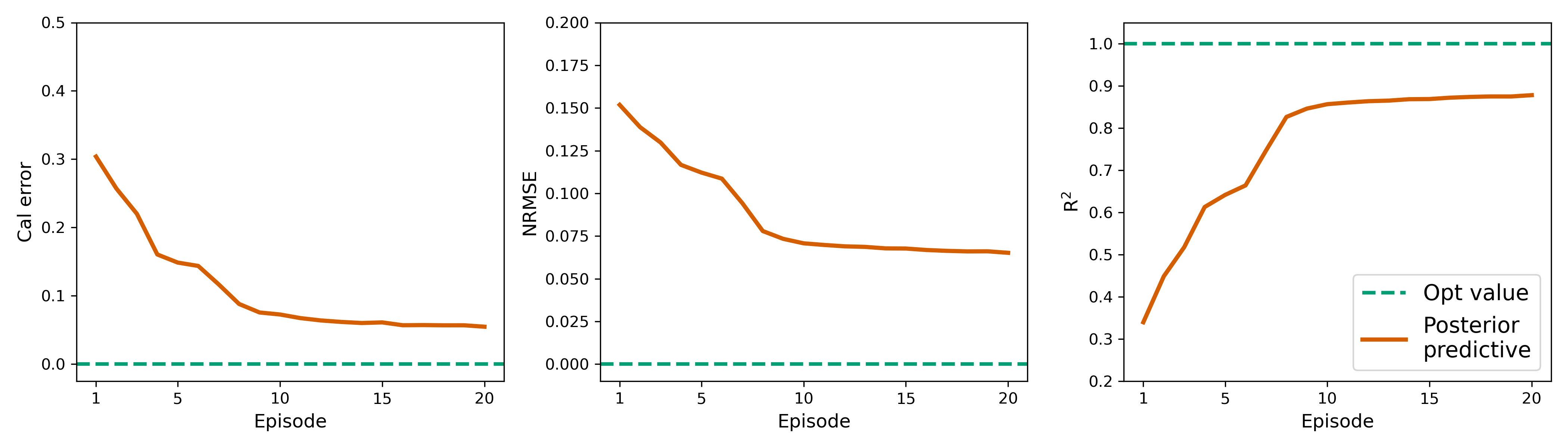}
    \caption{Data assimilation for DGFM on the European subset of the WeatherBench dataset. We infer the prequential posteriors after each episode ($\sim 1$ year) and plot the diagnostics computed on a separate test data ($\sim 4$ years). The orange curve shows the value of the 
    the calibration metrics (calibration error, NRMSE, $R^2$) of the posterior predictives obtained after each assimilation episode and the predictive performance of the posteriors improve steadily over each training episode. The green dotted line represents the maximum achievable value for each metric. Steady decay of the calibration error and NRMSE curves, along with the upward trend of the $R^2$ curve indicates effective data assimilation via posterior parameter updates in DGFM.}
    \label{ChapPPfig:wb-cal}
\end{figure}

\begin{figure}[htbp]
    \centering
    \includegraphics[width=0.9\linewidth]{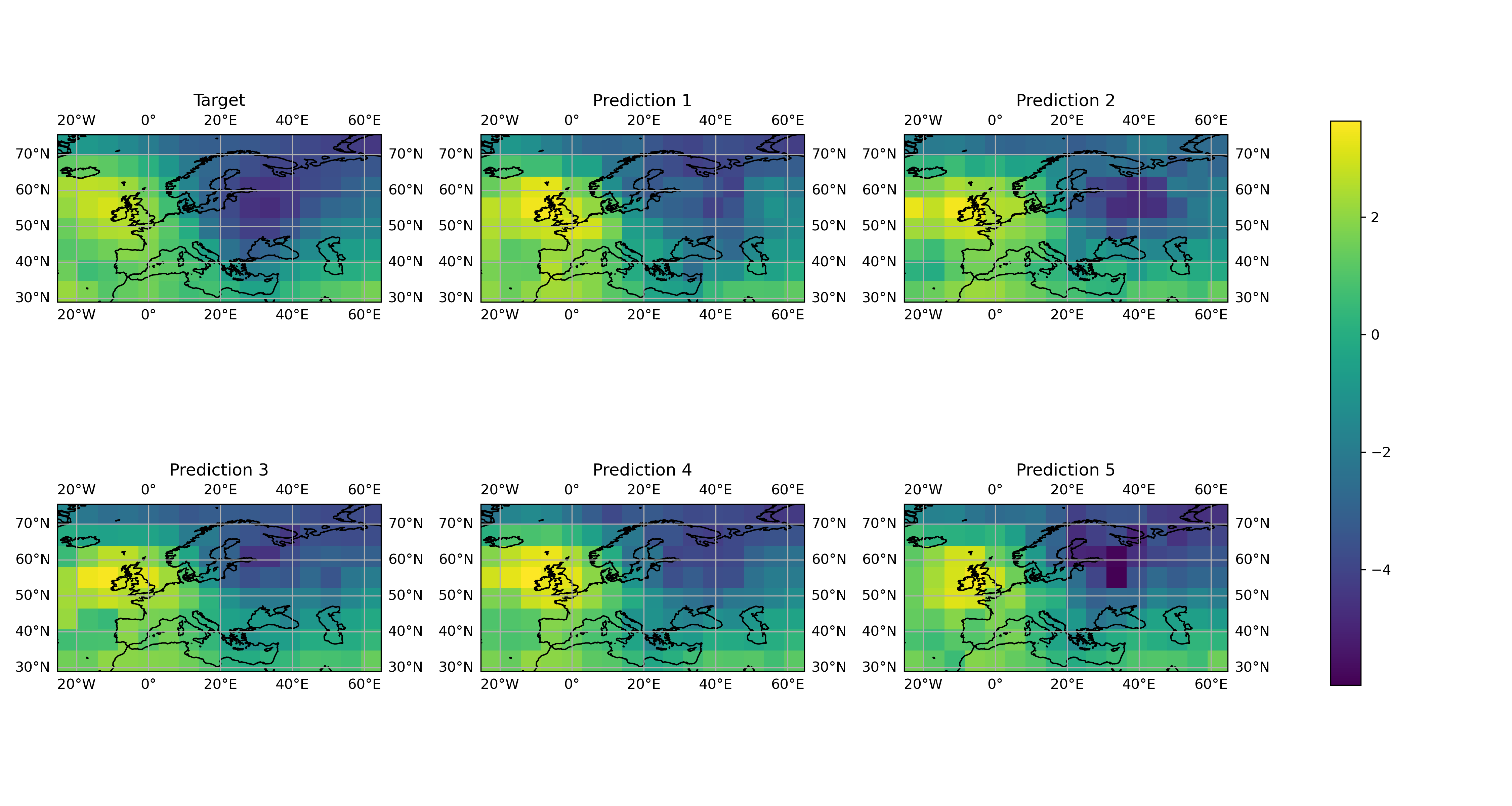}
    \caption{Comparison of target and simulations for the WeatherBench dataset. The top left image shows an observation from the year $2000$, while the remaining five images are simulated from the posterior predictive distributions generated from the final posterior predictive distribution. The simulations roughly capture the underlying spatial pattern.}
    \label{ChapPPfig:wb-prediction}
\end{figure}

For the WeatherBench experiment, we again infer prequential posteriors after each observation episode and evaluate predictive skill on the test set using calibration error, NRMSE, and $R^2$. As shown in Figure \ref{ChapPPfig:wb-cal}, assimilating additional episodes leads to a steady reduction in calibration error and normalized RMSE, while $R^2$ rises toward unity, demonstrating consistent improvements in forecast accuracy and calibration over the posterior updates. However, a more complex generative model could potentially achieve optimal performance. Our results show that the current model reaches its best predictive accuracy with sufficient training data. In Figure~\ref{ChapPPfig:wb-prediction}, we present sample predictions from the final posterior predictive distribution for a test-set observation. The simulated fields closely reproduce the underlying spatial structure.


\section{Conclusion}\label{ChapPPsec: conclusion}

In this work, we proposed a scalable and likelihood-free Bayesian framework for data assimilation, with a focus on improving probabilistic forecasting as new data becomes available. This framework can be used for complex implicit generative models like deep generative forecasting models (DGFMs). To address the challenge posed by intractable likelihoods, we adopted a generalized Bayesian approach based on the prequential loss, which allows for posterior updating without requiring explicit access to a likelihood function. This framework is particularly suitable for sequential learning and forecasting in temporally dependent settings. We also adopted an alternative notion of consistency, originally proposed by Dawid; which is appropriate under model misspecification. Rather than assuming the existence of a true parameter, we focus on identifying parameters that minimize predictive risk, thus aligning model training with the ultimate goal of forecasting accuracy. For posterior sampling, we implemented a waste-free sequential Monte Carlo (SMC) algorithm, augmented with a stochastic gradient MCMC (SGMCMC)–style forward kernel. To further enhance efficiency in high-dimensional settings, we employed adaptive preconditioning inspired by optimization techniques such as RMSprop. The resulting approach is scalable, parallelizable, and well-suited for training complex models like neural networks. However, note that the proposed sampler is an approximate algorithm: it targets the prequential posterior via a finite-particle SMC scheme with a biased 
Markov kernel. Therefore it does not inherit the exact theoretical guarantees of Section~\ref{ChapPPsec: gen bayes} in a strict sense. Nevertheless, the empirical results demonstrate that these approximations incur no meaningful loss in predictive performance across all settings considered.

We demonstrated the effectiveness of our approach on both a nonlinear time series model (Lorenz 96) and a meteorological image forecasting task (WeatherBench), highlighting its potential for real-world forecasting applications such as digital twin systems \cite{sharma2022digital}. For Lorenz 96, we compared our method against an ensemble Kalman filter with a commonly used misspecified state-space model for the system. We showed that our approach progressively learns an effective representation of the underlying dynamics, ultimately achieving superior predictive accuracy. This illustrates that when the true state-space model is unknown, the Kalman filter struggles, whereas our method can still perform considerably well. More broadly, in settings where an appropriate state-space model is unavailable or computationally prohibitive, our framework offers a practical model-updating mechanism. Its predictive strength follows from the universal approximation property of deep neural networks, which ensures that with sufficient data, the learned simulator can approximate the relevant physical relationships and deliver reliable forecasts.

We established asymptotic consistency of the prequential posterior under suitable regularity conditions on the simulator model. Verifying these conditions for flexible model classes such as deep generative models remains an important direction for future research. To the best of our knowledge, \cite{JMLR:v22:20-469} provides the most advanced theoretical treatment of generalized Bayes currently available, and our analysis of the prequential posterior builds on those results. When a likelihood function is available, however, there exist stronger Bernstein–von Mises results \cite{bhattacharya2019bayesian, bochkina2019bernstein}. Adapting those more refined techniques to the prequential setting could potentially yield sharper asymptotic guarantees, which we leave as future work. Other promising future directions include a more rigorous treatment of the approximation error introduced by the forward Markov kernel, as well as an analysis of the error bounds associated with the waste-free SMC procedure. Additionally, while we have used Gaussian and Student-\(t\) priors for the neural network parameters, exploring structured or shrinkage priors may further improve performance by introducing beneficial regularization in overparameterized settings.

\section{Conflict of interest}
On behalf of all authors, we state that there is no conflict of interest.

\bibliography{paper.bib}

\appendix


\section{Sketch proof of results related to the BvM theorem}

This section contains the complete statements of the lemmas and corollaries used in the main text to establish the Bernstein--von Mises (BvM) theorem, together with their corresponding proofs.

\subsection{Full statement of Lemma \ref{ChapPPlem-ullnm} }\label{ChapPPappendix-lemma1}

\textbf{[Martingale uniform law of large numbers]} Let us consider the assumptions, 
    \begin{enumerate}[label=\textbf{U\arabic*}]
    \setcounter{enumi}{0}
        \item \label{ChapPPu1}The metric space $(\Theta, d)$ is compact.
        \item \label{ChapPPu2} Suppose, $B(\theta, \rho)$ denotes a ball around $\theta$ of radius $\rho$ and define 
        $$
        \underline{\ell_t}(\theta, \rho) = \inf_{s \in B(\theta, \rho)} \ell_t(s) \quad \text{ and } \quad \overline{\ell_t}(\theta, \rho) = \sup_{s \in B(\theta, \rho)} \ell_t(s).
        $$

        For any $\theta \in \Theta$ there is $\rho(\theta)$ such that for all $\rho<\rho(\theta)$ the sequence of random variables $\left(\overline{\ell_t}(\theta, \rho)\right)$ and $\left(\underline{\ell_t}(\theta, \rho)\right)$ satisfy pointwise strong martingale LLN's (with common denominator an increasing predictable sequence $A_T$ ).
        \item \label{ChapPPu3}For all $\theta \in \Theta, {P}$-a.s.,
        $$
        \lim _{\rho \rightarrow 0} \limsup _{T \geq 1} \frac{1}{A_T} \sum_{t=1}^T E_{t-1}\left\{\bar{\ell}_t(\theta, \rho)-\underline{\ell_t}(\theta, \rho)\right\}=0.
        $$

    \end{enumerate}

\begin{unnumberedlemma}
Suppose the assumptions \ref{ChapPPu1}-\ref{ChapPPu3} holds, then $P$-a.s., 
\begin{equation*}
    \sup_{\theta \in \Theta} \mid \frac{1}{A_T} \sum_{t=1}^T[l_t(\theta) - E_{t-1} l_t(\theta)]\mid \rightarrow 0. 
\end{equation*}
\end{unnumberedlemma}

Often, the assumptions \ref{ChapPPu2}, \ref{ChapPPu3} are difficult to verify. However, assumption \ref{ChapPPassumption2} implies \ref{ChapPPu2} and assumption \ref{ChapPPassumption3} implies \ref{ChapPPu3}, which are relatively easier to check.

\subsection{Proof of Corollary \ref{ChapPPcor}}
\label{ChapPPappendix-cor}
\begin{unnumberedcorollary}
Under the assumptions \ref{ChapPPassumption1}-\ref{ChapPPassumption4}, it follows from Lemma \ref{ChapPPlem-ullnm} that,
    $$
    \sup_{\theta \in \Theta} |\bar{L}_T(\theta) - \bar{L}^*(\theta)| \to 0 \text{ as } T \to \infty,
    $$
    with probability one under $P$.
\end{unnumberedcorollary}
\begin{proof}
    Under assumption \ref{ChapPPassumption4}, for a fixed $\epsilon >0$ there exists a $T_1(\epsilon)$ such that for all $T > T_1(\epsilon)$ with probability $1$ under $P$, 
\begin{align}\label{ChapPPeqn-sup1}
    &|\bar{L}_T^*(\theta) - \bar{L}^*(\theta)| < \epsilon/2, \text{ for all } \theta \in \Theta \notag\\
    &\implies \sup_{\theta \in \Theta} |\bar{L}_T^*(\theta) - \bar{L}^*(\theta)| < \epsilon/2.
\end{align}
Similarly, under assumptions \ref{ChapPPassumption1}-\ref{ChapPPassumption3}, Lemma \ref{ChapPPlem-ullnm} implies that for the fixed $\epsilon >0$ there exists a $T_2(\epsilon)$ such that for all $T > T_2(\epsilon)$ with probability $1$ under $P$, 
\begin{equation}\label{ChapPPeqn-sup2}
    \sup_{\theta \in \Theta} | \bar{L}_T(\theta) - \bar{L}_T^*(\theta)| < \epsilon/2.
\end{equation}
On combining equations \ref{ChapPPeqn-sup1} and \ref{ChapPPeqn-sup2}, for any $T > \max \{T_1(\epsilon), T_2(\epsilon)\}$ 
\begin{align}\label{ChapPPeqn-ulln2}
    \sup_{\theta \in \Theta} |\bar{L}_T(\theta) - \bar{L}^*(\theta)| &\leq \sup_{\theta \in \Theta} \left\{|\bar{L}_T(\theta) - \bar{L}_T^*(\theta)| +  |\bar{L}^*_T(\theta) - \bar{L}^*(\theta)|\right\} \notag\\
    &< \epsilon/2 + \epsilon/2 = \epsilon,
\end{align}
with probability $1$ under $P$.
Since equation \ref{ChapPPeqn-ulln2} holds for an arbitrary $\epsilon >0$, it follows that, with probability $1$ under $P$,
$$
\sup_{\theta \in \Theta} |\bar{L}_T(\theta) - \bar{L}^*(\theta)| \to 0 \text{ as } T \to \infty.
$$

\end{proof}

\subsection{Lemma on the existence of an extremum estimator}\label{ChapPPlem-existence}

\begin{unnumberedlemma}
Let $(\Omega, \mathcal{F})$ be a measurable space, and let $(\Theta, d)$ be a compact, separable space. Let $Q: \Omega \times \Theta \to \mathbb{R} \cup \{-\infty, \infty\}$ be such that $Q(\cdot, \theta)$ is $\mathcal{F}$-measurable for each $\theta$ in $\Theta$, and $Q(\omega, \cdot)$ is continuous for all $\omega$ in an event $F \in \mathcal{F}$. Then, there exists a function $\hat{\theta}: \Omega \to \Theta$ such that $\hat{\theta}$ is $\mathcal{F}$-measurable and for all $\omega$ in $F$
\begin{equation*}
    Q(\omega, \hat{\theta}(\omega)) = \inf_{\theta \in \Theta} Q(\omega, \theta).
\end{equation*}
    
\end{unnumberedlemma}

\subsection{Proof of Lemma \ref{ChapPPlem-consistency}}
\label{ChapPPappendix-lem-consistent_estimators}
\begin{unnumberedlemma}
    (Asymptotic consistency) Under assumptions \ref{ChapPPassumption1}-\ref{ChapPPassumption7}, as $T \to \infty$, we have with probability $1$ under $P$,
    $$
    d(\hat{\theta}_T, \theta^*) \to 0.
    $$
\end{unnumberedlemma}
We present here a proof of the lemma based on Theorem 5.1 in \cite{skouras1998optimal} as adapted by \cite{pacchiardi2024probabilistic}.

\begin{proof}
From assumption \ref{ChapPPassumption7}, for a fixed $\epsilon >0$ it is possible to find a $\delta(\epsilon) > 0 $ such that, 
\begin{equation}\label{ChapPPeqn-delta11}
    \min_{\theta: d(\theta, \theta^*) \geq \epsilon} \bar{L}^*(\theta) - \bar{L}^*(\theta^*) = \delta(\epsilon),
\end{equation}
with probability $1$ under $P$.

Due to Corollary \ref{ChapPPcor}, with probability $1$ under $P$, there exists a $T_1(\delta(\epsilon))$ such that for all $T > T_1(\delta(\epsilon))$
\begin{equation*}
    |\bar{L}_T(\theta^*) - \bar{L}^*(\theta^*)| < \delta(\epsilon)/2,
\end{equation*}
which implies
\begin{align}\label{ChapPPeqn-delta1}
    \bar{L}^*(\theta^*) &> \bar{L}_T(\theta^*) - \delta(\epsilon)/2 \notag\\
    & \geq \bar{L}_T(\hat{\theta}_T) - \delta(\epsilon)/2,
\end{align}
where the second inequality is from the definition of $\hat{\theta}_T$.

On exploiting Corollary \ref{ChapPPcor} once again, we can define a $T_2(\delta(\epsilon))$ such that for all $T > T_2(\delta(\epsilon))$
\begin{equation}\label{ChapPPeqn-delta2}
    | \bar{L}_T(\hat{\theta}_T) - \bar{L}^*(\hat{\theta}_T)| < \delta(\epsilon)/2,
\end{equation} 
with probability $1$ under $P$.

Then, with probability $1$ under $P$, for all $T > \max\{T_1(\delta(\epsilon)), T_2(\delta(\epsilon))\}$ 
\begin{align}\label{ChapPPeqn-delta3}
    \bar{L}^*(\hat{\theta}_T) - \bar{L}^*(\theta^*) &= \bar{L}^*(\hat{\theta}_T) - \bar{L}_T(\hat{\theta}_T) + \bar{L}_T(\hat{\theta}_T) - \bar{L}^*(\theta^*)\notag \\
    & < \delta(\epsilon)/2 + \delta(\epsilon)/2 = \delta(\epsilon)
\end{align}
from equation \ref{ChapPPeqn-delta1} and equation \ref{ChapPPeqn-delta2}.

Note that equation \ref{ChapPPeqn-delta3} ensures that the difference considered in equation \ref{ChapPPeqn-delta11} is smaller than $\delta(\epsilon)$ when $\theta = \hat{\theta}_T$. However, by equation \ref{ChapPPeqn-delta11}, the same difference is at least $\delta(\epsilon)$ for all $\theta$ that are outside the $\epsilon$-radius ball around $\theta^*$. This implies that, $\hat{\theta}_T$ must lie inside the $\epsilon$-radius ball, meaning that $d(\hat{\theta}_T, \theta^*) < \epsilon$ with probability $1$ under $P$. Since this is true for any $\epsilon > 0$, it follows that, with probability $1$ under $P$
$$
d(\hat{\theta}_T, \theta^*) \to 0\text{ as } T \to \infty.
$$
   
\end{proof}

\subsection{Sketch proof of Theorem \ref{ChapPPth-bvm}}\label{ChapPPappendix-bvm}

\begin{proof}
From Lemma \ref{ChapPPlem-consistency}, we have, with probability one under $P$, as $T \to \infty$, 
$$
d(\hat{\theta}_T, \theta^*) \to 0.
$$
which proves the existence of a sequence of estimators $\{\hat{\theta}_T\}_{T=1}^{\infty}$ such that as $T \to \infty$, with probability one under $P$, $d(\hat{\theta}_T, \theta^*) \to 0$ under the assumptions \ref{ChapPPassumption1}-\ref{ChapPPassumption7}.

By Theorem 6 from \cite{JMLR:v22:20-469}, using the above sequence of estimators $\{\hat{\theta}_T\}_{T=1}^{\infty}$ along with the assumptions \ref{ChapPPassumption9}-\ref{ChapPPassumption11}, $L_T(\theta)$ can be represented as,
\begin{equation}\label{ChapPPeq-taylorseries}
    \bar{L}_T(\theta) = \bar{L}_T(\hat{\theta}_T) + \frac{1}{2} (\theta - \hat{\theta}_T)'H_T (\theta - \hat{\theta}_T) + r_T(\theta - \hat{\theta}_T)
\end{equation}

where $H_T = \bar{L}_T''(\hat{\theta}_T) \in \mathbb{R}^{p \times p}$ is symmetric and $H_T \to  H^*$. There exists $\epsilon_0, c_0 >0$ such that, for all $T$ sufficiently large, for all $\theta \in B_{\epsilon_0}(\mathbf{0})$, we have $|r_T(\theta)| \leq c_0|\theta|^3. $ 


Finally, on using the Taylor series expansion of the function $\bar{L}_T(\theta)$ in equation \ref{ChapPPeq-taylorseries}, together with the assumptions \ref{ChapPPassumption8} and \ref{ChapPPassumption12}, the Theorem \ref{ChapPPth-bvm} holds (by Theorem 4 from \cite{JMLR:v22:20-469}).

For completeness, Theorems~4 and~6 from \cite{JMLR:v22:20-469} are restated below as Theorems~\ref{ChapPPth-miller4} and~\ref{ChapPPth-miller6}, respectively.

\end{proof}

\begin{theorem}\label{ChapPPth-miller4}
[Adapted from Theorem 4 from \cite{JMLR:v22:20-469}]
Fix $\theta^* \in \mathbb{R}^p$ and let $\pi: \mathbb{R}^p \rightarrow \mathbb{R}$ be a probability density with respect to Lebesgue measure such that $\pi$ is continuous at $\theta^*$ and $\pi\left(\theta^*\right)>0$. Let $\bar{L}_T: \mathbb{R}^p \rightarrow \mathbb{R}$ for $T \in \mathbb{N}$ and assume:

\begin{enumerate}[label=\textbf{M\arabic*}]
    \setcounter{enumi}{0}
    \item\label{ChapPPassumption-miller1} $\bar{L}_T$ can be represented as
$$
\bar{L}_T(\theta) = \bar{L}_T(\hat{\theta}_T) + \frac{1}{2} (\theta - \hat{\theta}_T)'H_T (\theta - \hat{\theta}_T) + r_T(\theta - \hat{\theta}_T)
$$
where $\hat{\theta}_T \in \mathbb{R}^p$ such that $\hat{\theta}_T \rightarrow \theta^*, H_T \in \mathbb{R}^{p \times p}$ symmetric such that $H_T \rightarrow H^*$ for some positive definite $H^*$, and $r_T: \mathbb{R}^p \rightarrow \mathbb{R}$ has the following property: there exist $\varepsilon_0, c_0>0$ such that for all $T$ sufficiently large, for all $\theta \in B_{\varepsilon_0}(\mathbf{0})$, we have $\left|r_T(\theta)\right| \leq c_0|\theta|^3;$

\item For any $\varepsilon>0, \liminf _T \inf _{\theta \in B_{\varepsilon}\left(\hat{\theta}_T\right)^c}\left(\bar{L}_T(\theta)-\bar{L}_T\left(\hat{\theta}_T\right)\right)>0$,

then defining $z_T = \int_{\Theta} \exp(-{L}_T(\theta)) \pi(\theta)d\theta$ and $\pi_{P}(\theta|y^{T}) = \pi(\theta) \exp(-{L}_T(\theta))/ z_T$ we have,
    \begin{equation*}
        \int_{B_{\epsilon}(\theta^*)} \pi_{P}(\theta|y^{T}) d\theta \underset{T \rightarrow \infty}{\longrightarrow} 1 \quad \text{for all } \epsilon >0,
    \end{equation*}
    which means, $\pi_{P}(\theta|y^{T})$ concentrates at $\theta^*$;
    \begin{equation*}
        z_T \approx \frac{\exp(-{L}_T(\hat{\theta}_T)) \pi(\theta^*)}{\mid \det H^*\mid^{1/2}} \left(\frac{2\pi}{A_T}\right)^{p/2}
    \end{equation*}
    as $T \to \infty$ (Laplace approximation), and letting $q_T$ be the density of $\sqrt{A_T}(\theta - \hat{\theta}_T)$ when $\theta \sim \pi_{GP}(\theta|y^{T})$,
    \begin{equation*}
        \int_{\Theta}\left | q_T(\theta) - \mathcal{N}(\theta \mid \mathbf{0}, {H^*}^{-1}) \right| d\theta \underset{T \rightarrow \infty}{\longrightarrow} 0,
    \end{equation*}
    which implies, $q_T$ converges to $\mathcal{N}(\mathbf{0}, {H^*}^{-1})$ in total variation.
\end{enumerate}

\end{theorem}
Note that, we use $A_T$ instead of $T$ as in Theorem 4 from \cite{JMLR:v22:20-469}. Since any $A_T$ satisfying Assumption~\ref{ChapPPassumption2} is an increasing function of $T$, Theorem~\ref{ChapPPth-miller4} continues to hold for such choices of $A_T$. 
This connection is crucial for extending the Bernstein-von Mises result in our setting, 
and we therefore apply Theorem~\ref{ChapPPth-miller4} to ultimately derive Theorem~\ref{ChapPPth-bvm}.

\begin{theorem}\label{ChapPPth-miller6}
[Adapted from Theorem 6 from \cite{JMLR:v22:20-469}]Let $E \subseteq \mathbb{R}^p$ be open and convex, and let $\theta^* \in E$. Let $\bar{L}_T: E \rightarrow \mathbb{R}$ have continuous third derivatives, and assume:

\begin{enumerate}[label=\textbf{M\arabic*}]
    \setcounter{enumi}{2}
\item there exist $\hat{\theta}_T\in E$ such that $\hat{\theta}_T \rightarrow \theta^*$ and $\bar{L}_T^{\prime}(\hat{\theta}_T)=0$ for all $T$ sufficiently large,
\item $\bar{L}_T^{\prime \prime}\left(\theta^*\right) \rightarrow H^*$ as $T \rightarrow \infty$ for some positive definite $H^*$, and
\item $\bar{L}_T^{\prime \prime \prime}$ is uniformly bounded;

\end{enumerate}

Then, letting $H_T=\bar{L}_T^{\prime \prime}(\hat{\theta}_T)$, condition \ref{ChapPPassumption-miller1} of Theorem \ref{ChapPPth-miller4} is satisfied for all $T$ sufficiently large.

\end{theorem}

\section{Derivation of the coefficients of the stationary AR$(5)$ process}\label{ChapPPappendix: stationary coefficients ar5}
We generate data from the autoregressive process

\begin{align*}
    Y_t &= \theta^{*\prime} Y_{t-q:t-1} + \epsilon_t,
    \qquad
    \epsilon_t \sim \mathcal{N}(0,1), \\
    \theta^* &= [\theta_1^*, \dots, \theta_q^*]'.
\end{align*}

To ensure stationarity of the process, the parameter vector \(\theta^*\) is constructed through the partial autocorrelation parameterization using the Levinson--Durbin recursion \citep{levinson1947,durbin1960,monahan1984}. Specifically, partial autocorrelation coefficients (PACF) are independently generated as

\[
\phi_{kk}
\overset{\mathrm{iid}}{\sim}
\mathrm{Uniform}(-0.9,0.9),
\qquad
k=1,\dots,q,
\]

and recursively transformed into autoregressive coefficients according to

\begin{align*}
\delta_j^{(k)}
&=
\delta_j^{(k-1)}
-
\phi_{kk}
\delta_{k-j}^{(k-1)},
\qquad
j=1,\dots,k-1,\\
\delta_k^{(k)}
&=
\phi_{kk}.
\end{align*}

The final autoregressive parameter vector is then defined as

\[
\theta^*
=
[
\delta_1^{(q)},
\dots,
\delta_q^{(q)}
]'.
\]

Since each partial autocorrelation satisfies \(|\phi_{kk}|<1\), the resulting autoregressive process is stationary by construction \citep{monahan1984}.

For the experiment reported here, this procedure yields

\[
\theta^*
=
[
0.6689,\,
0.0401,\,
-0.4582,\,
0.9265,\,
-0.1972
]'.
\]

\section{Verification of regularity assumptions for AR(q) process}\label{ChapPPappendix-assumption_check}
First, we list some inequalities used to check the regularity assumptions \ref{ChapPPassumption2}-\ref{ChapPPassumption4}. Then, we verify those for a stationary AR(q) process as considered in Section~\ref{ChapPPsec: arq_process}.

\subsection{Some important inequalities}

\subsubsection{Proof of the Convexity Power Inequality}\label{ChapPPproof-convexity_inequality}

For any convex function $f(x)$ and non-negative inputs $a, b \geqslant 0$, Jensen's inequality gives us
\[
f\left(\frac{a+b}{2}\right) \leqslant \frac{f(a) + f(b) }{2}
\]
Since $f(x) = x^\gamma$ is strictly convex for any power $\gamma > 1$, substituting it into the expression yields:
\begin{align*}
    \left(\frac{a+b}{2}\right)^\gamma &\leqslant \frac{a^\gamma + b^\gamma}{2}\\
    \implies (a+b)^\gamma &\leqslant 2^{\gamma-1}\left(a^\gamma + b^\gamma\right)
\end{align*}

\subsubsection{Derivation of the mean value theorem (MVT) bound}\label{ChapPPmvt_bound}

 
Consider the function $f(x) = x^\beta$ defined on $x \geqslant 0$. Since $\beta > 1$, $f(x)$ is continuously differentiable with derivative:
\[
f'(x) = \beta x^{\beta-1}
\]
By the Mean Value Theorem, for any $a, b \geqslant 0$, there exists a point $c$ strictly between $a$ and $b$ such that:
\[
f(a) - f(b) = f'(c)(a - b)
\]
Taking the absolute value of both sides yields:
\[
\left| a^\beta - b^\beta \right| = \beta c^{\beta-1} |a - b|
\]
Since $c$ lies between $a$ and $b$, and the function $x^{\beta-1}$ is strictly increasing for $\beta > 1$:
\[
c^{\beta-1} \leqslant a^{\beta-1} + b^{\beta-1}
\]
Combining these steps yields the final functional inequality used in the derivation:
\[
\left| a^\beta - b^\beta \right| \leqslant \beta |a - b| \left( a^{\beta-1} + b^{\beta-1} \right)
\]

\subsection{Verifying \ref{ChapPPassumption2}}

\ref{ChapPPassumption2}: With ${P}$ probability $1$, there exists an $A_t < \infty$ for all $t=1, 2, \cdots$ such that one of the following conditions hold,
    \begin{enumerate}
        \item $$
    \sum_{t=1}^{\infty} \frac{\mathbb{E}_{t-1}\left\{\sup _{\theta \in \Theta}\left|\ell_t(\theta)\right|^2\right\}}{A_t^2}<\infty; \text{ or,}
    $$

    \item there exists an $\epsilon >0$ such that
    $$
    \left(\sum_{t=1}^T \mathbb{E}_{t-1} \{\sup_{\theta \in \Theta} |\ell_t(\theta)|^2\}\right)^{(1+ \epsilon)/2} = \mathcal{O}(A_T).
    $$
    \end{enumerate}

\begin{proof}
Here we verify the 1st condition.
\begin{align*}
    l_t(\theta) &= 2~\mathbb{E}_Q \left| (\theta - \theta^*)' Y_{t-q : t-1} + \omega_t -\epsilon_t  \right|^\beta - \mathbb{E}_Q |\omega_t - \omega_t'|^{\beta}\\
    & = 2~\mathbb{E}_Q \left|  (\theta - \theta^*)' Y_{t-q : t-1} + \omega^{(1)}_t  \right|^\beta - C_{\eta, \beta},
\end{align*}
where $0 \leqslant C_{\eta, \beta} = \mathbb{E} |Z_1|^{\beta} < \infty$ as $Z_1 \sim \mathcal{N}(0, 2)$ and $\omega^{(1)}_t = \omega_t  -\epsilon_t \sim \mathcal{N}(0,2)$. Then, 

\begin{align}\label{ChapPPeqn:assumption2check}
\mathbb{E}_{t-1} &\left[ \sup_{\theta \in \Theta} |l_t(\theta)|^2 \right]\notag \\[0.5em]
&= \mathbb{E}_{t-1} \left[ \sup_{\theta \in \Theta} \left( 2~\mathbb{E}_Q \left|  (\theta - \theta^*)' Y_{t-q : t-1} + \omega^{(1)}_t \right|^\beta - C_{\eta,\beta} \right)^2 \right]\notag \\[0.5em]
&\leqslant   \mathbb{E}_{t-1} \left[ \sup_{\theta \in \Theta} \left( 2~\mathbb{E}_Q \left|  (\theta - \theta^*)' Y_{t-q : t-1} + \omega^{(1)}_t \right|^\beta \right)^2 \right] + C_{\eta,\beta}^2,
\end{align}
since $(a-b)^2 \leqslant a^2 + b^2$ for $a, b \geqslant 0$. Next, we consider two cases where $\beta \in (0, 1]$ and $\beta \in (1, 2)$.

\paragraph{Case 1: $\beta \in (0, 1]$}
\begin{align*}
\mathbb{E}_{t-1} &\left[ \sup_{\theta \in \Theta} \left( 2~\mathbb{E}_Q \left|  (\theta - \theta^*)' Y_{t-q : t-1} + \omega^{(1)}_t \right|^\beta \right)^2 \right] \\[0.5em]
&\leqslant 4 \, \mathbb{E}_{t-1} \left[ \sup_{\theta \in \Theta} \left( \mathbb{E}_Q \left[ \left|  (\theta - \theta^*)' Y_{t-q : t-1} \right|^\beta + |\omega^{(1)}_t|^\beta \right] \right)^2 \right]  \tag{by subadditivity for $\beta \in (0, 1]$ meaning that $ |a+b|^{\beta} \leqslant ||a| + |b||^{\beta} \leqslant |a|^{\beta} + |b|^{\beta}$ for $a, b \in \mathbb{R}$} \\[0.5em]
&\leqslant 8\, \mathbb{E}_{t-1} \left[ \sup_{\theta \in \Theta} \left( \mathbb{E}_Q \left|  (\theta - \theta^*)' Y_{t-q : t-1} \right|^\beta\right)^2 + \left(\mathbb{E}_Q|\omega^{(1)}_t|^\beta\right)^2  \right]  \tag{since $(a-b)\geqslant 0 \implies$ $2ab \leqslant a^2 + b^2 \implies (a+b)^2 \leqslant 2(a^2+b^2)$ for  $a, b \geqslant 0$} \\[0.5em]
&= 8\, \mathbb{E}_{t-1} \left[ \sup_{\theta \in \Theta} \left| (\theta - \theta^*)' Y_{t-q : t-1} \right|^{2\beta} + C_{\eta, \beta}^2  \right]   \\[0.5em]
&\leqslant 8\,\mathbb{E}_{t-1} \left[ \sup_{\theta \in \Theta}  (\|\theta - \theta^*\|\| Y_{t-q : t-1}\| )^{2\beta} \right] + 8 \, C_{\eta, \beta}^2   \tag{using Cauchy-Schwarz inequality} \\[0.5em]
&= 8\, (b_{\theta}(\theta^*))^{2\beta} \mathbb{E}_{t-1} \| Y_{t-q : t-1}\| ^{2\beta}  + 8\,C_{\eta, \beta}^2    \tag{where $\sup_{\theta \in \Theta}  \|\theta - \theta^*\| = b_{\theta}(\theta^*) < \infty$ since $\Theta$ is compact} \\
\implies \mathbb{E}_{t-1} & \left[ \sup_{\theta \in \Theta} |l_t(\theta)|^2 \right] \leqslant 8 \,(b_{\theta}(\theta^*))^{2\beta}  \| y_{t-q : t-1}\| ^{2\beta} + 9\,C_{\eta, \beta}^2  = c_{t, \beta}.\tag{from Equation~\ref{ChapPPeqn:assumption2check}}
\end{align*}

Now let us consider the inequality
\begin{align*}
\|Y_{t-q:t-1}\|^{2\beta}
&\le
q^{\max(\beta-1,0)}
\sum_{i=1}^{q}|Y_{t-i}|^{2\beta}.
\end{align*}

Since \(\{Y_t\}\) is a stationary AR process with i.i.d.\ Gaussian innovations,

\[
\mathbb{E}|Y_t|^{2\beta}
=
\mathbb{E}|Y_0|^{2\beta}
=
\mu(2\beta)
<
\infty,
\qquad
\forall t.
\]

Therefore,

\begin{align*}
\mathbb{E}\|Y_{t-q:t-1}\|^{2\beta}
&\le
q^{\max(\beta-1,0)}
\sum_{i=1}^{q}\mathbb{E}|Y_{t-i}|^{2\beta}
\\
&=
q^{\max(\beta-1,0)}\,
q\,\mu(2\beta)
\\
&=
q^{\max(\beta,1)}\mu(2\beta)
<
\infty.
\end{align*}

Hence,

\begin{align*}
\sum_{t=q+1}^{\infty}
q^{\max(\beta,1)}
\frac{\mu(2\beta)}{t^2}
&<
\infty
\\
\Rightarrow\quad
\sum_{t=q+1}^{\infty}
\frac{
\mathbb{E}\|Y_{t-q:t-1}\|^{2\beta}
}{t^2}
&<
\infty
\\
\Rightarrow\quad
\mathbb{E}
\left[
\sum_{t=q+1}^{\infty}
\frac{
\|Y_{t-q:t-1}\|^{2\beta}
}{t^2}
\right]
&<
\infty
\tag{swapping the expectation and the infinite summation by monotone convergence theorem}
\\
\Rightarrow\quad
\sum_{t=q+1}^{\infty}
\frac{
\|Y_{t-q:t-1}\|^{2\beta}
}{t^2}
&<
\infty,
\qquad
P\text{-a.s.}
\\
\Rightarrow\quad
\sum_{t=q+1}^{\infty}
\frac{
8\bigl(b_0(\theta^*)\bigr)^{2\beta}
\|Y_{t-q:t-1}\|^{2\beta}
+
9\, C_{\eta,\beta}^{2}
}{t^2}
&<
\infty \qquad
P\text{-a.s.}
\\
\Rightarrow\quad
\sum_{t=q+1}^{\infty} \frac{\mathbb{E}_{t-1}\left\{\sup _{\theta \in \Theta}\left|\ell_t(\theta)\right|^2\right\}}{A_t^2}
&<
\infty \qquad
P\text{-a.s., taking } A_t = t.
\end{align*}


\paragraph{Case 2: $ \beta \in (1,2)$}

\begin{align*}
\mathbb{E}_{t-1} &\left[ \sup_{\theta \in \Theta} \left( 2~\mathbb{E}_Q \left|  (\theta - \theta^*)' Y_{t-q : t-1} + \omega^{(1)}_t \right|^\beta \right)^2 \right] \\[0.5em]
&\leqslant 4 \, \mathbb{E}_{t-1} \left[ \sup_{\theta \in \Theta} \left( \mathbb{E}_Q  \left( \left| (\theta - \theta^*)' Y_{t-q : t-1} \right| + |\omega^{(1)}_t| \right)^{\beta}  \right)^2 \right] \tag{by triangle inequality} \\[0.5em]
&\leqslant 4 \, \mathbb{E}_{t-1} \left[ \sup_{\theta \in \Theta} \mathbb{E}_Q  \left( \left|  (\theta - \theta^*)' Y_{t-q : t-1} \right| + |\omega^{(1)}_t| \right)^{2\beta}  \right]  \tag{by Jensen's inequality} \\[0.5em]
&\leqslant 4 \cdot 2^{2\beta-1} \, \mathbb{E}_{t-1} \left[ \sup_{\theta \in \Theta} \mathbb{E}_Q \left[ \left| (\theta - \theta^*)' Y_{t-q : t-1} \right|^{2\beta} + |\omega^{(1)}_t|^{2\beta}  \right] \right] \tag{via convexity with $\gamma = 2\beta >2$, see Section~\ref{ChapPPproof-convexity_inequality}} \\[0.5em]
&=  2^{2\beta+1} \left[\mathbb{E}_{t-1} \left( \sup_{\theta \in \Theta}  \left| (\theta - \theta^*)' Y_{t-q : t-1} \right|^{2\beta} \right)+  \mathbb{E}_Q|\omega^{(1)}_t|^{2\beta}   \right] \\[0.5em]
&\leqslant 2^{2\beta +1} \, \left[ ( b_{\theta}(\theta^*))^{2\beta}\| y_{t-q : t-1} \|^{2\beta}  +  C_{\eta, 2\beta}\right] \tag{where $\sup_{\theta \in \Theta}  \|\theta - \theta^*\| = b_{\theta}(\theta^*) < \infty$ since $\Theta$ is compact, same as Case 1} \\[0.5em]
\implies \mathbb{E}_{t-1} & \left[ \sup_{\theta \in \Theta} |l_t(\theta)|^2 \right] \leqslant 2^{2\beta +1} \, \left[ ( b_{\theta}(\theta^*))^{2\beta}\| y_{t-q : t-1} \|^{2\beta}  +  C_{\eta, 2\beta}\right] +  C_{\eta, \beta}^2  = c_{t, \beta}< \infty\tag{from Equation~\ref{ChapPPeqn:assumption2check}}
\end{align*}
Then, on taking $A_t = t$, similarly as Case 1, we get, 

\begin{align*}
    \sum_{t=q+1}^{\infty} \frac{\mathbb{E}_{t-1}\left\{\sup _{\theta \in \Theta}\left|\ell_t(\theta)\right|^2\right\}}{A_t^2} < \infty.
\end{align*}

\end{proof}

\subsection{Verifying \ref{ChapPPassumption3}}
\ref{ChapPPassumption3}: For each $\theta \in \Theta$, there is a constant $\tau>0$, such that for every $s, d(\theta, s) \leq \tau$ implies that for every $t \geq 1$, ${P}$-a.s.,
    $$
    \left|\ell_t(\theta)-\ell_t(s)\right| \leq B_t\left(Y_t\right) h\{d(\theta, s)\}
    $$
    where $\left\{B_t(\cdot)\right\}$ is a sequence of $\mathcal{F}_{t}$-measurable functions such that ${P}$-a.s.,
    $$
    \limsup _T \frac{1}{A_T} \sum_{t=1}^T \mathbb{E}_{t-1}\left\{B_t\left(Y_t\right)\right\}<\infty
    $$
    and $h(\cdot)$ is a non-random function such that $h(y) \downarrow h(0)=0$ as $y \downarrow 0$. The null sets, $B_t(\cdot)$ and $h$ may depend on $\theta$.

\begin{proof}

\begin{align}\label{ChapPPeqn:assumption3check}
|l_t(\theta) - l_t(s)| 
&= 2~\left| \mathbb{E}_Q\left[| \theta' Y_{t-q : t-1} - Y_t + \omega_t |^{\beta}\right] - \mathbb{E}_Q\left[| s' Y_{t-q : t-1} - Y_t + \omega_t |^{\beta}\right] \right|\notag\\[0.5em]
&\leqslant 2~\mathbb{E}_Q \left| | \theta' Y_{t-q : t-1} - Y_t + \omega_t |^{\beta} - | s' Y_{t-q : t-1} - Y_t + \omega_t |^{\beta} \right|,
\end{align}
since $|\mathbb{E}[X]| \leqslant \mathbb{E}|X|$ for a random variable $X$. Now let us consider
two cases for $\beta \in (0, 1]$ and $\beta \in (1, 2)$.

\paragraph{Case 1: $\beta \in (0, 1]$}
\begin{align*}
\mathbb{E}_Q &\left| | \theta' Y_{t-q : t-1} - Y_t + \omega_t |^{\beta} - | s' Y_{t-q : t-1} - Y_t + \omega_t |^{\beta} \right|\\
&\leqslant \mathbb{E}_Q \left[ \left| |\theta' Y_{t-q : t-1} - Y_t + \omega_t| - |s' Y_{t-q : t-1} - Y_t + \omega_t| \right|^\beta \right]
\tag{by subadditivity for $\beta \leqslant 1$}\\[0.5em]
&[\text{Since, for any $a, b \geqslant 0, |a+b|^{\beta} \leqslant a^{\beta} + b^{\beta}$ for $\beta \in (0, 1]$. On assuming $|x|\geqslant|y|$} \\ 
&\text{without loss of generality; taking $a = |x|-|y|, b = |y|$ we get the above inequality. }]\\
&\leqslant \mathbb{E}_Q \left[ \left| (\theta' Y_{t-q : t-1} - Y_t + \omega_t) - (s' Y_{t-q : t-1} - Y_t + \omega_t) \right|^\beta \right]
\tag{by Reverse Triangle Inequality} \\[0.5em]
&= \mathbb{E}_Q \left[ \left| (\theta - s)' Y_{t-q : t-1} \right|^\beta \right] 
\tag{noise terms cancel completely} \\[0.5em]
&\leqslant \mathbb{E}_Q \left[ \left( \|\theta - s\| \cdot \|Y_{t-q : t-1}\| \right)^\beta \right]
\tag{by Cauchy-Schwarz inequality} \\[0.5em]
&= \|\theta - s\|^\beta \cdot \mathbb{E}_Q \left[ \|Y_{t-q : t-1}\|^\beta \right] \\[0.5em]
\implies |l_t(\theta) - l_t(s)| 
&\leqslant 2~\|\theta - s\|^\beta \cdot \mathbb{E}_Q \left[ \|Y_{t-q : t-1}\|^\beta \right] = h(\|\theta - s\|) \cdot B_t(Y_{1:t}) \tag{from Equation \ref{ChapPPeqn:assumption3check}}
\end{align*}
\noindent where $h(x) = 2x^\beta$ and $B_t(Y_{1:t}) = \mathbb{E}_Q \left[ \|Y_{t-q : t-1}\|^\beta \right] = \|Y_{t-q : t-1}\|^\beta = c_t $. 

Now let us consider the inequality
\begin{align}\label{ChapPPeqn:bounding_expected_ytbeta}
\mathbb{E}\|Y_{t-q:t-1}\|^{\beta}
&\le
\mathbb{E} (q^{\max(\beta/2-1,0)}
\sum_{i=1}^{q}|Y_{t-i}|^{\beta})\notag\\
&=
 q^{\max(\beta/2-1,0)}
\sum_{i=1}^{q} \mathbb{E}|Y_{t-i}|^{\beta}\notag\\
&=
 q^{\max(\beta/2,1)} \mu(\beta) < \infty,
\end{align}

The true data-generating process is a stationary $\mathrm{AR}(q)$,
\[
Y_t \;=\; \sum_{i=1}^{q} \theta_i^*\, Y_{t-i} + \epsilon_t,
\qquad \epsilon_t \overset{\mathrm{iid}}{\sim} \mathcal{N}(0,1),
\]
with the roots of the characteristic polynomial lying strictly outside the
unit disc, ensuring stationarity. Hence the process is ergodic by Theorem 14.23 from \cite{hansen2022econometrics}. Next, we state a lemma adapted from Proposition 1 of \cite{betken2021ordinal}, which follows from Birkhoff's ergodic theorem.
\begin{lemma}\label{ChapPPlemma_birkhoff_betken}
Suppose that $\{Y_t\}_t$ is a stationary ergodic process, and that $g: \mathbb{R}^q \to \mathbb R$ is a measurable function such that $\mathbb E |g(Y_{t-q: t-1})| < \infty$. Then, 
\begin{equation}
    \frac1T\sum_{t=q+1}^{T} g (Y_{t-q: t-1}) \to \mathbb E g (Y_{1: q}),
\end{equation}
almost surely, as $T \to \infty$.   
\end{lemma}
We apply the above lemma to the following, on setting $A_T = T$,



\begin{align*}
    \limsup _T \frac{1}{A_T} \sum_{t=q+1}^T\mathbb{E}_{t-1} B_t(Y_{1:t}) &= \limsup _T \frac{1}{T} \sum_{t=q+1}^T c_t\\
    &= \limsup _T \frac{1}{T} \sum_{t=q+1}^T \|y_{t-q:t-1}\|^{\beta}\\
    & = \mathbb{E}\|Y_{t-q:t-1}\|^{\beta} 
     < \infty, \, P-\text{almost surely.}\tag{from Equation~\ref{ChapPPeqn:bounding_expected_ytbeta} and using Lemma~\ref{ChapPPlemma_birkhoff_betken}.}
\end{align*}

\paragraph{Case 2: $\beta \in (1, 2)$}

\begin{align*} 
\mathbb{E}_Q &\left| | \theta' Y_{t-q : t-1} - Y_t + \omega_t |^{\beta} - | s' Y_{t-q : t-1} - Y_t + \omega_t |^{\beta} \right| \\[0.5em]
&\leqslant \mathbb{E}_Q \left[ \beta \left| (\theta - s)' Y_{t-q : t-1} \right| \cdot \left( |\theta' Y_{t-q : t-1} - Y_t + \omega_t |^{\beta-1} + |s' Y_{t-q : t-1} - Y_t + \omega_t |^{\beta-1} \right) \right] 
\tag{by mean value bound for $\beta > 1$; see Section~\ref{ChapPPmvt_bound} } \\[0.5em]
&\leqslant \beta \|\theta - s\| \cdot \mathbb{E}_Q \left[ \|Y_{t-q : t-1}\| \cdot \left( |\theta' Y_{t-q : t-1} - Y_t + \omega_t |^{\beta-1} + |s' Y_{t-q : t-1} - Y_t + \omega_t |^{\beta-1} \right) \right]
\tag{by Cauchy-Schwarz inequality} \\[0.5em]
\implies |l_t(\theta) - l_t(s)| 
&\leqslant  h(\|\theta - s\|) \cdot B_t(Y_{1:t})\tag{from Equation \ref{ChapPPeqn:assumption3check}}
\end{align*}
\noindent where $h(x) = 2\beta x$ and
\begin{equation}\label{ChapPP_appendix_B_t_arq}
 B_t(Y_{1:t}) = \mathbb{E}_Q \left[ \|Y_{t-q : t-1}\| \cdot \left( |\theta' Y_{t-q : t-1} - Y_t + \omega_t |^{\beta-1} + |s' Y_{t-q : t-1} - Y_t + \omega_t |^{\beta-1} \right) \right].
\end{equation}
 Then, 
\begin{align}\label{ChapPPeqn:expectation_Bt}
    \mathbb{E}_{t-1} B_t(Y_{1:t}) &= \|y_{t-q : t-1}\| \mathbb{E}_Q \left( |\theta' y_{t-q : t-1} - y_t + \omega_t |^{\beta-1} + |s' y_{t-q : t-1} - Y_t + \omega_t |^{\beta-1} \right)\notag\\
    & = \|y_{t-q : t-1}\| \mathbb{E}_Q \left( Z_t(\theta)^{\beta - 1} + Z_t(s)^{\beta - 1}\right),
\end{align}
where $Z_t(\theta) = \sum_{i=1}^q (\theta_i - \theta^*_i)y_{t-i} + \omega_t - \epsilon_t$. $Z_t(\theta)$ is distributed as a Gaussian random variable with 

\begin{align*}
    \mathbb{E}_Q Z_t(\theta) = \sum_{i=1}^q (\theta_i - \theta^*_i)y_{t-i},  \qquad \mathbb{V}_Q Z_t(\theta)= 2.
\end{align*}
Therefore, 

\begin{equation}
    \limsup _T \frac{1}{T} \sum_{t=q+1}^T\mathbb{E}_{t-1} B_t(Y_{1:t}) < \infty, \, P-\text{a.s.} \tag{similarly as Case 1}
\end{equation}

\end{proof}

\subsection{Verifying \ref{ChapPPassumption4}}
\ref{ChapPPassumption4}: There exists a function \( \bar{L}^*(\theta) \) such that as \( T \to \infty \), \( \bar{L}_T^*(\theta) \to \bar{L}^*(\theta) \) uniformly with probability one under $P$.
\begin{proof}

Recall from the verification of \ref{ChapPPassumption2} above that $A_T=T$ for this example, so throughout this proof $\bar{L}_T^*(\theta) = \frac1T\sum_{t=q+1}^T \mathbb{E}_{t-1}\ell_t(\theta)$.

\begin{align*}
\mathbb{E}_{t-1}\,\ell_t(\theta)
&=
\mathbb{E}_{t-1}
\left[
2\,\mathbb{E}_{Q}
\left|
(\theta-\theta^{*})' Y_{t-q:t-1}
+\omega_t - \epsilon_t
\right|^{\beta}
-
\mathbb{E}_{Q}|\omega_t - \omega_t'|^{\beta}
\right]
\\
&=
2\,\mathbb{E}_{Q}
\left|
\mu_t(\theta) + Z_1
\right|^{\beta}-\mathbb{E}_{Q}|Z_1|^{\beta}\\
& =:\; h_\beta(\mu_t(\theta)),
\end{align*}
where $Z_1 \sim \mathcal{N}(0,2)$ and $\mu_t(\theta) = (\theta-\theta^{*})'y_{t-q:t-1}$. We provide the full expression of the function $h_\beta(\mu_t(\theta))$ in Section~\ref{ChapPP_appendix_h_beta_function}.

Since $\{Y_t\}_t$ is a stationary Gaussian AR$(q)$ process, for
any fixed $\theta$, $\mu_t(\theta)=(\theta-\theta^*)'y_{t-q:t-1}$ is Gaussian and therefore has,
\[
\mathbb{E}\left|\mu_t(\theta)\right|^\beta<\infty
\qquad\text{for every }\beta \in (0,2).
\]
Moreover, since $Z_1\sim\mathcal{N}(0,2)$, we have $\mathbb{E}_Q|Z_1|^\beta<\infty$ for $\beta \in (0,2)$.

Using the inequality
\[
|a+b|^\beta
\leq C_\beta\left(|a|^\beta+|b|^\beta\right),
\qquad a,b\in\mathbb{R},
\]
for some finite constant $C_\beta>0$, we obtain for a fixed $\theta$ for every $\beta \in (0,2)$,
\[
\begin{aligned}
\mathbb{E}\left[
\left|h_\beta\bigl(\mu_t(\theta)\bigr)\right|
\right]
&\leq
\mathbb{E}_Q\left|\mu_t(\theta)+Z_1\right|^\beta
+
\mathbb{E}_Q|Z_1|^\beta\\
&\leq
C_\beta\mathbb{E}\left|\mu_t(\theta)\right|^\beta
+
(C_\beta+1)\mathbb{E}_Q|Z_1|^\beta\\
&<\infty.
\end{aligned}
\]
Next, we apply Birkhoff's ergodic theorem \emph{directly to $\{h_\beta(\mu_t(\theta))\}_t$} (via Lemma~\ref{ChapPPlemma_birkhoff_betken}) and we get for every fixed $\theta\in\Theta$, $P$-almost surely,
\begin{equation}
\bar{L}_T^*(\theta) = \frac1T\sum_{t=q+1}^{T} h_\beta\big(\mu_t(\theta)\big)
\;\xrightarrow{\mathrm{a.s.}}\;
\mathbb{E}\big[h_\beta(\mu_{q+1}(\theta))\big] =: \bar{L}^*(\theta), \qquad T\to\infty, \label{ChapPPeqn:C4-pointwise}
\end{equation}
where the expectation is taken with $(Y_1,\ldots,Y_q)$ distributed according to the marginal law of the stationary process.

Now, recall from the verification of \ref{ChapPPassumption3} that there are a non-random function $h(\cdot)$ with $h(y)\downarrow h(0)=0$ as $y\downarrow0$, and $\mathcal F_{t-1}$-measurable functions $B_t(\cdot)$ (given explicitly there, separately for $\beta\in(0,1]$ and $\beta\in(1,2)$), such that $P$-a.s., for every $t\ge q+1$ and every $d(\theta,s)\le\tau$,
\[
|\ell_t(\theta)-\ell_t(s)| \le B_t(Y_{1:t})\,h(d(\theta,s)).
\]
Taking $\mathbb{E}_{t-1}[\cdot]$, using $|\mathbb E_{t-1}X|\le \mathbb E_{t-1}|X|$, and writing $\bar{B}_t := \mathbb{E}_{t-1}\{B_t(Y_{1:t})\}$, this passes to the conditional risk, $P$-a.s. we get,
\begin{equation}
\big|h_\beta(\mu_t(\theta)) - h_\beta(\mu_t(s))\big|
= \big|\mathbb{E}_{t-1}\ell_t(\theta)-\mathbb{E}_{t-1}\ell_t(s)\big|
\le \bar{B}_t\, h(d(\theta,s)). \label{ChapPPeqn:C4-lipschitz}
\end{equation}
As in the verification of \ref{ChapPPassumption3}, $\bar B_t$ is a fixed function of the observation window $Y_{t-q:t-1}$ (equal to $\|Y_{t-q:t-1}\|^\beta$ for $\beta\in(0,1]$; and to the corresponding expression in $Z_t(\theta)^{\beta-1}$ for $\beta\in(1,2)$ as given in Equation~\eqref{ChapPPeqn:expectation_Bt}) and has finite mean. Hence, we again apply Lemma~\ref{ChapPPlemma_birkhoff_betken} and get (by taking $g = B_t: \mathbb R^q \to \mathbb R$) $P$-a.s.,
\begin{equation}\label{ChapPP_appendix_limitBt}
    \frac1T\sum_{t=q+1}^T \bar{B}_t \;\xrightarrow{\mathrm{a.s.}}\; \mathbb{E}[\bar{B}_{q+1}] =: \bar{B} < \infty.
\end{equation}
Averaging Equation~\eqref{ChapPPeqn:C4-lipschitz} over $t=q+1,\dots,T$, for all $d(\theta,s)\le\tau$, $P$-a.s. we have,
\begin{equation}\label{ChapPP_appendix_limit_Lt_thetas}
\big|\bar L_T^*(\theta)-\bar L_T^*(s)\big| \le \frac{1}{T}\sum_{t=q+1}^T\big|\mathbb{E}_{t-1}\ell_t(\theta)-\mathbb{E}_{t-1}\ell_t(s)\big| \le \Big(\tfrac1T\textstyle\sum_{t=q+1}^T \bar B_t\Big) h(d(\theta,s)).
\end{equation}
Letting $T\to\infty$ termwise and using Equation~\eqref{ChapPPeqn:C4-pointwise}, for all $d(\theta,s)\le\tau$, $P$-a.s. we get,
\begin{equation}\label{ChapPPeqn:C4-equicont}
    \big|\bar L^*(\theta)-\bar L^*(s)\big| \le \bar B\, h(d(\theta,s)). 
\end{equation}

Next, let us fix $\varepsilon>0$.
\begin{enumerate}[label=(\roman*)]
\item \emph{Choose a mesh size.} Since $h(y)\downarrow h(0)=0$ as $y\downarrow0$, pick $\delta\in(0,\tau]$ small enough that
\[
(\bar B+1)\,h(\delta) < \varepsilon/3.
\]
\item \emph{Cover $\Theta$.} Since $(\Theta,d)$ is compact (\ref{ChapPPassumption1}), it admits a finite cover by $\delta$-balls centred at $\theta_1,\dots,\theta_{N(\delta)}$.
\item \emph{Intersect finitely many a.s.\ events.} The pointwise convergence in \eqref{ChapPPeqn:C4-pointwise} applied at each of the $N(\delta)$ centres, together with Equation~\eqref{ChapPP_appendix_limitBt}, we have $P$-a.s.,
\[
\bar L_T^*(\theta_j) \to \bar L^*(\theta_j) \ \ (j=1,\dots,N(\delta)),
\qquad
\frac1T\sum_{t=q+1}^T \bar B_t \to \bar B.
\]

\item \emph{Find a path-dependent $T_0$.} Choose $T_0$ (depending on the sample path) such that, for all $T\ge T_0$, $P$-a.s.,
\[
\big|\bar L_T^*(\theta_j)-\bar L^*(\theta_j)\big| < \varepsilon/3 \ \ (j=1,\dots,N(\delta)),
\qquad
\frac1T\sum_{t=q+1}^T \bar B_t \le \bar B + 1.
\]
\item \emph{Bound $\big|\bar L_T^*(\theta)-\bar L^*(\theta)\big|$ uniformly.} Let $\theta\in\Theta$ be arbitrary, and pick a centre $\theta_j$ with $d(\theta,\theta_j)\le\delta$. By the triangle inequality,
\[
\big|\bar L_T^*(\theta)-\bar L^*(\theta)\big|
\;\le\;
\big|\bar L_T^*(\theta)-\bar L_T^*(\theta_j)\big|
\;+\;
\big|\bar L_T^*(\theta_j)-\bar L^*(\theta_j)\big|
\;+\;
\big|\bar L^*(\theta_j)-\bar L^*(\theta)\big|,
\]
and for every $T\ge T_0$, each term on the right is smaller than $\varepsilon/3$ with probability $1$ under $P$:
\begin{itemize}
\item $\big|\bar L_T^*(\theta)-\bar L_T^*(\theta_j)\big| \le \big(\tfrac1T\sum_{t=q+1}^T \bar B_t\big) h(\delta) \le (\bar B+1)h(\delta) < \varepsilon/3$, by Equation~\eqref{ChapPP_appendix_limit_Lt_thetas} and step (iv);
\item $\big|\bar L_T^*(\theta_j)-\bar L^*(\theta_j)\big| < \varepsilon/3$, directly by step (iv);
\item $\big|\bar L^*(\theta_j)-\bar L^*(\theta)\big| \le \bar B\, h(\delta) \le (\bar B+1)h(\delta) < \varepsilon/3$, by Equation~\eqref{ChapPPeqn:C4-equicont} and step (i).
\end{itemize}
Summing the three bounds, $\big|\bar L_T^*(\theta)-\bar L^*(\theta)\big| < \varepsilon$ for every $T\ge T_0$, $P$-a.s. Crucially, $\delta$ (hence the cover, hence $T_0$) was fixed before $\theta$ was chosen, so $T_0$ does not depend on $\theta$.
\end{enumerate}
Since $\varepsilon>0$ was arbitrary, we have shown that with probability $1$ under $P$,
\[
\sup_{\theta\in\Theta} \big|\bar L_T^*(\theta)-\bar L^*(\theta)\big| \;\longrightarrow\; 0 \qquad \text{as } T\to\infty,
\]
i.e.\ $\bar L_T^*(\theta)\to\bar L^*(\theta)$ uniformly over $\Theta$, $P$-almost surely, which is precisely \ref{ChapPPassumption4}.

\end{proof}

\subsubsection{Full expression of $h_\beta(\mu_t(\theta))$}\label{ChapPP_appendix_h_beta_function}
\begin{align*}
\mathbb{E}_{t-1}\,\ell_t(\theta)
&=
\mathbb{E}_{t-1}
\left[
2\,\mathbb{E}_{Q}
\left|
(\theta-\theta^{*})' Y_{t-q:t-1}
+\omega_t - \epsilon_t
\right|^{\beta}
-
\mathbb{E}_{Q}|\omega_t - \omega_t'|^{\beta}
\right]
\\
&=
2\,\mathbb{E}_{Q}
\left|
\mu_t(\theta) + Z_1
\right|^{\beta}
-
\mathbb{E}_{Q}|Z_1|^{\beta}\\
& =:\; h_\beta(\mu_t(\theta)),
\end{align*}
where $Z_1 \sim \mathcal{N}(0,2)$ and $\mu_t(\theta) = (\theta-\theta^{*})'y_{t-q:t-1}$.

We now use the identity for the absolute moment \citep{winkelbauer2012moments} of a Gaussian random variable. if \(X \sim \mathcal{N}(\mu,\sigma^2)\), then for \(\beta > -1\),

\[
\mathbb{E}[|X|^\beta]
=
\frac{(2\sigma^2)^{\beta/2}}{\sqrt{\pi}}
\Gamma\left(\frac{\beta+1}{2}\right)
M\left(
-\frac{\beta}{2},
\frac{1}{2};
-\frac{\mu^2}{2\sigma^2}
\right),
\]

where \(M(a,b;z)\) denotes Kummer's confluent hypergeometric function \citep{magnus1966formulas} of the first kind, defined as

\[
M(a,b;z)
=
\sum_{k=0}^{\infty}
\frac{(a)_k}{(b)_k}
\frac{z^k}{k!},
\]

where \((a)_k\) denotes the Pochhammer symbol (rising factorial), given by

\[
(a)_k =
\begin{cases}
1, & k = 0, \\
a(a+1)\cdots(a+k-1), & k > 0.
\end{cases}
\]

Applying this with $\sigma^2 = 2$ (as $Z_1 \sim \mathcal{N}(0,2)$):
\begin{align*}
\mathbb{E}_Q|\mu + Z_1|^{\beta}
&=
2^{\beta/2}\,
\frac{2^{\beta/2}\,\Gamma\!\left(\tfrac{\beta+1}{2}\right)}{\sqrt{\pi}}\;
M\!\left(-\tfrac{\beta}{2};\,\tfrac{1}{2};\,-\tfrac{\mu^2}{4}\right)
\\
&=
\frac{2^{\beta}\,\Gamma\!\left(\tfrac{\beta+1}{2}\right)}{\sqrt{\pi}}\;
M\!\left(-\tfrac{\beta}{2};\,\tfrac{1}{2};\,-\tfrac{\mu^2}{4}\right),
\end{align*}
and at $\mu = 0$:
\[
\mathbb{E}_Q|Z_1|^{\beta}
=
\frac{2^{\beta}\,\Gamma\!\left(\tfrac{\beta+1}{2}\right)}{\sqrt{\pi}},
\]
since $M(-\beta/2;\,1/2;\,0) = 1$.
Therefore,
\begin{align*}
\mathbb{E}_{t-1}\,\ell_t(\theta)
&=
\frac{2^{\beta+1}\,\Gamma\!\left(\tfrac{\beta+1}{2}\right)}{\sqrt{\pi}}
\left[
M\!\left(-\tfrac{\beta}{2};\,\tfrac{1}{2};\,-\tfrac{\mu_t(\theta)^2}{4}\right)
- \tfrac{1}{2}
\right]
\\
&=\; h_\beta(\mu_t(\theta)).
\end{align*}

\section{Sensitivity analysis of the tuning parameters for the AR($q$) process} \label{ChapPPappendix:sensitivity}

We first investigate the impact of retaining the Metropolis-Hastings (MH) correction in the preconditioned Markov kernel for different learning rates ($\eta$). We also provide the sensitivity analysis of the total count of particles ($N$) and the length of Markov chain ($P$) used within SMC for the AR($q$) process.

As described in Section~\ref{ChapPPsec: arq_process}, we fix an initial training dataset and sequentially update the posterior distribution upon observing batches of \(\tau = 10\) new observations. Following each update, we resample \(1000\) posterior samples from the SMC approximation and estimate the kernel Stein discrepancy (KSD) with respect to the target prequential posterior as a diagnostic measure of sample convergence using unbiased estimate of the gradient of the target. As before, predictive performance is evaluated using calibration error, normalized root mean squared error (NRMSE), and the coefficient of determination (\(R^2\)), all computed on a held-out test dataset of length \(300\).

\subsection{Impact of MH correction and sensitivity analysis of the learning rate in SMC}

\begin{figure}
    \centering
    \includegraphics[width=0.99\linewidth]{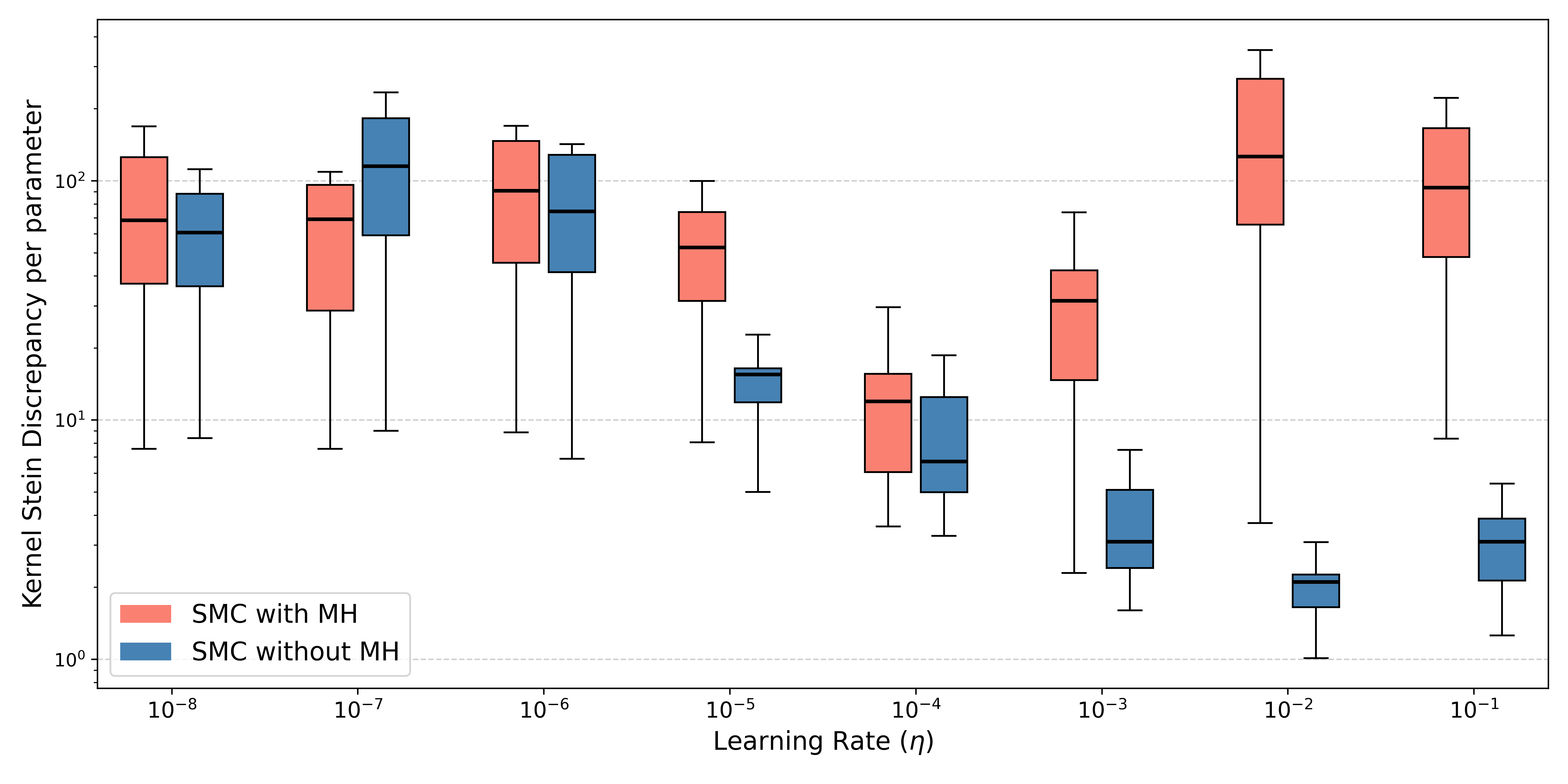}
    \caption{Comparison of posterior sample convergence under SMC with and without the Metropolis-Hastings (MH) correction in the preconditioned forward Markov kernel. Boxplots show the distribution of KSD values across $20$ episodes, and the black line denotes the median for different learning rates $(\eta)$, with $N=150$ and $P=10$.
}
    \label{ChapPPfig:mh_boxplot_ar5}
\end{figure}

\begin{figure}
    \centering
    \includegraphics[width=0.99\linewidth]{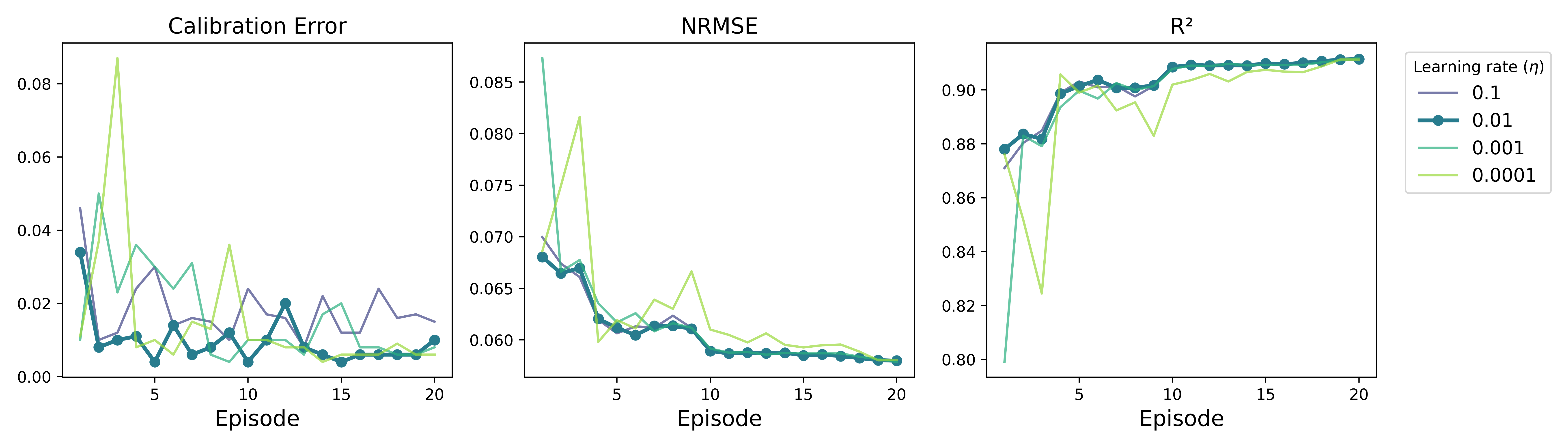}
    \caption{Predictive performance of the posterior samples obtained using different learning rates ($\eta$) and without the MH correction in SMC, where $N=150$, $P=10$. The curve corresponding to $(\eta = 0.01)$, which is used for the experiments reported in the main text, is highlighted. For clarity of presentation, results for \(\eta < 10^{-4}\) are omitted, as they exhibit significantly poorer diagnostic performance.
}
    \label{ChapPPfig:epsilon_cal_ar5}
\end{figure}

We compare the convergence of SMC samples to the target posterior in terms of the kernel Stein discrepancy (KSD), with and without the Metropolis-Hastings accept-reject step in the forward kernel used within SMC, as shown in Figure~\ref{ChapPPfig:mh_boxplot_ar5}. Here we have kept the particle counts (\(N\)) and the length of Markov chain ($P$) fixed at $150, 10$ respectively. The kernel without MH correction performs best for learning rates \(\eta \in [10^{-3}, 10^{-1}]\), whereas the kernel with correction suffers from a high rejection rate for large values of \(\eta\).

However, when \(\eta\) is very small (e.g., \(\eta < 10^{-5}\)), both samplers perform poorly, as the Markov chain exhibits negligible movement due to the extremely small step size. Overall, we conclude that the kernel without MH correction provides better mixing and is more robust to the choice of \(\eta\), provided that \(\eta\) is not too small.

Next, we analyze the predictive performance of the posterior samples obtained from SMC without the MH step, for different values of \(\eta\). All diagnostic measures share similar trends over the training episodes, as shown in Figure~\ref{ChapPPfig:epsilon_cal_ar5}.

\subsection{Sensitivity analysis of the total count of particles ($N$) in SMC}

\begin{figure}
    \centering
    \includegraphics[width=0.99\linewidth]{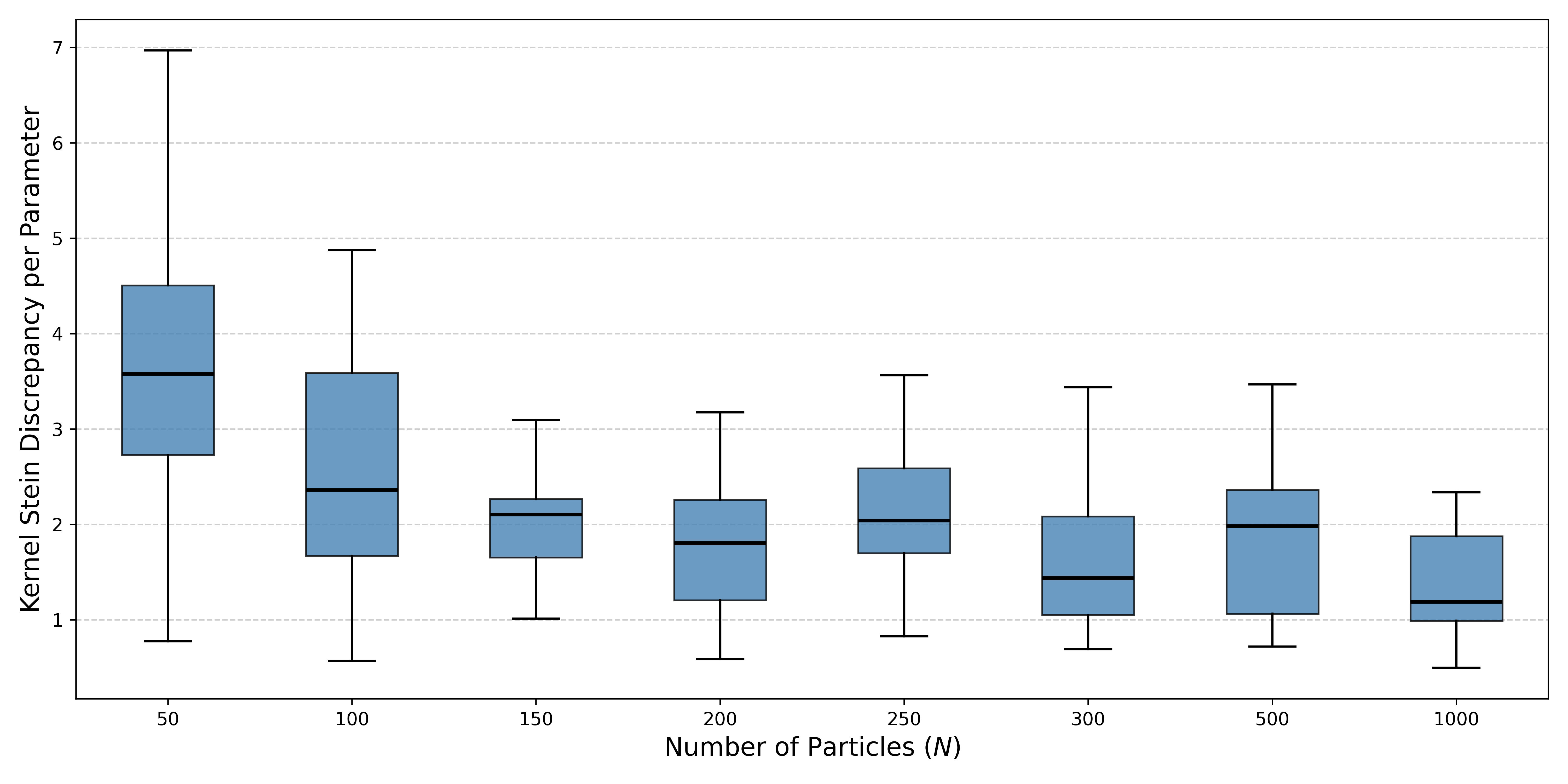}
    \caption{Comparison of posterior sample convergence under SMC across different particle counts \((N)\). Boxplots represent the distribution of KSD values over \(20\) episodes with $\eta = 0.01$ and $P=10$, and the black line denotes the median.
}
    \label{ChapPPfig:nparticle_boxplot_ar5}
\end{figure}

\begin{figure}
    \centering
    \includegraphics[width=0.99\linewidth]{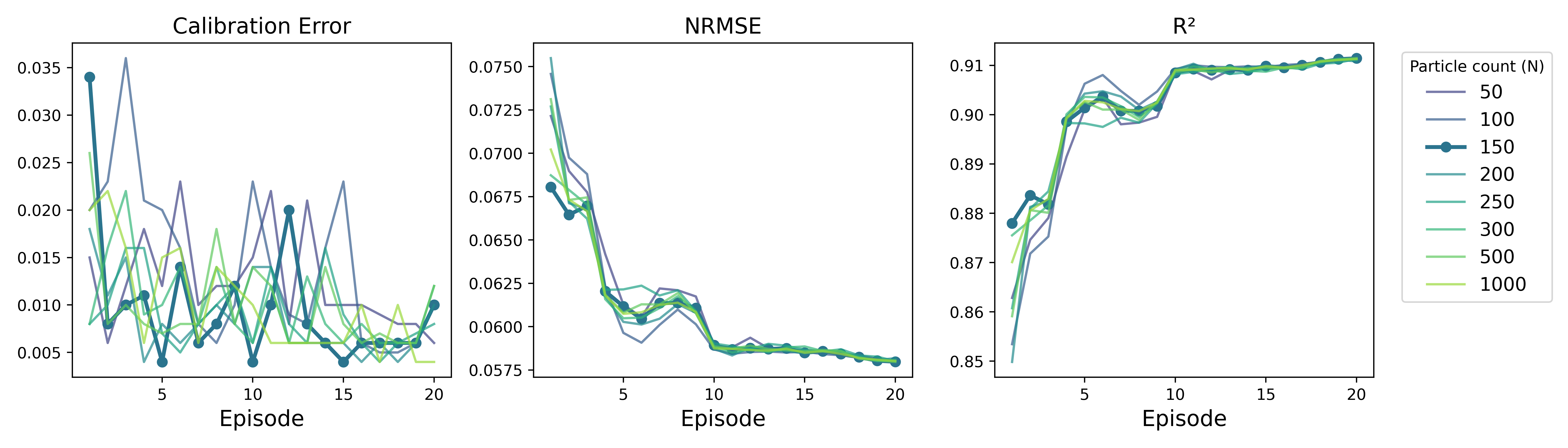}
    \caption{Predictive performance of the posterior samples obtained using different particle counts \((N)\) with $\eta = 0.01$ and $P=10$. The curve corresponding to $N = 150$, which is used for the experiments reported in the main text, is highlighted.}
    \label{ChapPPfig:nparticle_cal_ar5}
\end{figure}

We continue our experiments without the Metropolis-Hastings (MH) correction in the kernel used within SMC with $\eta = 0.01$ and $P=10$. In Figure~\ref{ChapPPfig:nparticle_boxplot_ar5}, we show the convergence of posterior samples obtained using different particle counts (\(N\)) in SMC. The discrepancy decreases uniformly as \(N\) increases; however, we do not observe significant improvements for values of \(N\) beyond \(150\), and therefore use this setting in the main text.

We also report the corresponding predictive performance in Figure~\ref{ChapPPfig:nparticle_cal_ar5}. The diagnostic curves show similar trends across training episodes, with performance improving as the number of episodes increases.

\subsection{Sensitivity analysis of the length of Markov chain ($P$) in (wastefree) SMC}

\begin{figure}
    \centering
    \includegraphics[width=0.99\linewidth]{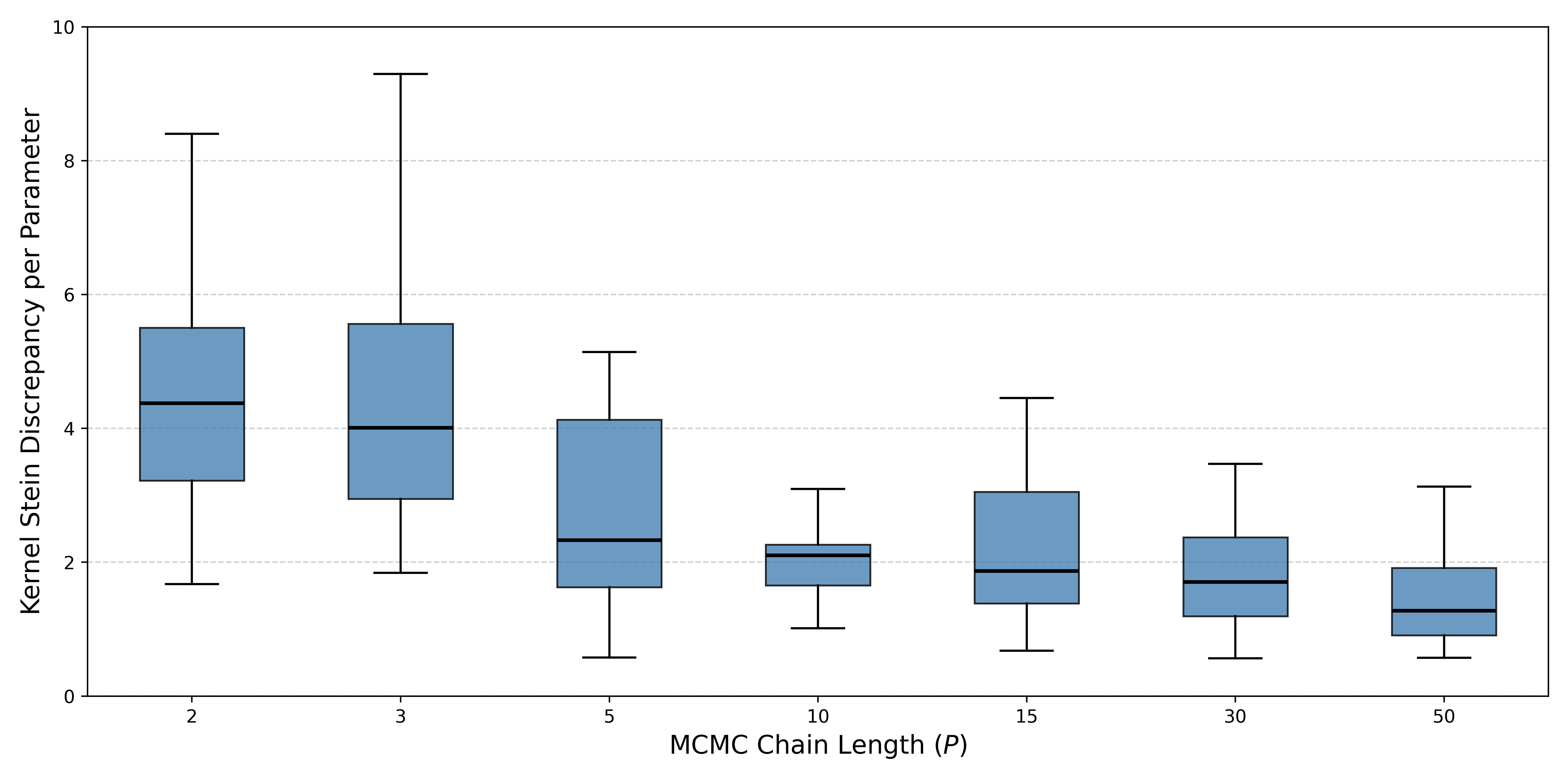}
    \caption{Comparison of posterior sample convergence under SMC across different lengths of the Markov chain \((P)\) in SMC with $\eta=0.01$ and $N=150$, meaning that $M = 75, ~50, ~30, ~15, ~10, ~5, ~3$ respectively. Boxplots represent the distribution of KSD values over \(20\) episodes, and the black line denotes the median.}
    \label{ChapPPfig:nmcmc_boxplot_ar5}
\end{figure}

\begin{figure}
    \centering
    \includegraphics[width=0.99\linewidth]{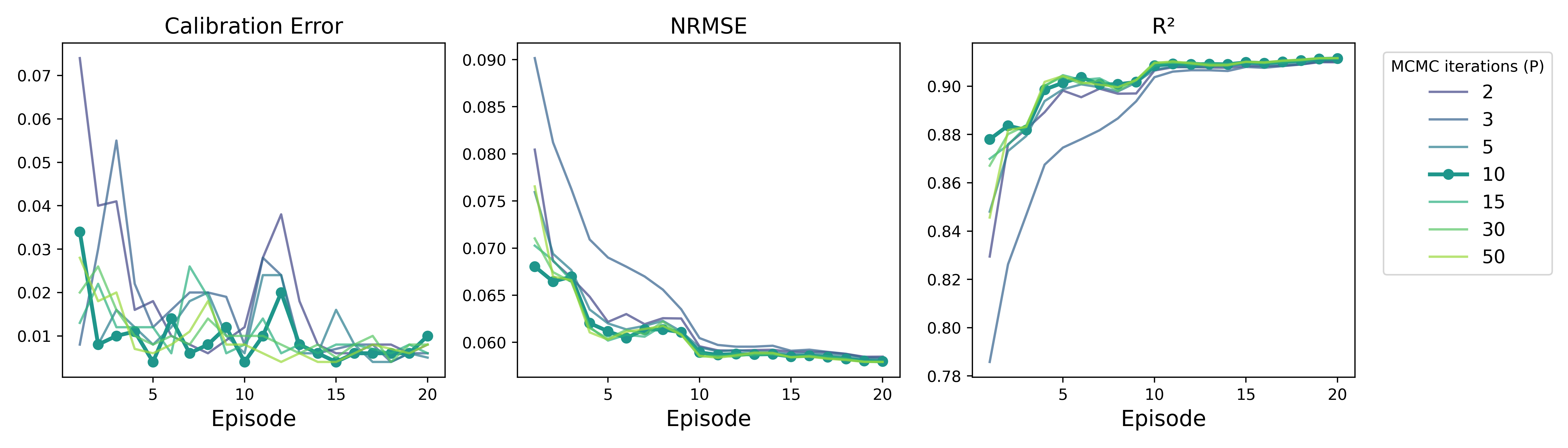}
    \caption{Predictive performance of the posterior samples obtained using different length of Markov chain \((P)\) with $\eta=0.01$ and $N=150$. The curve corresponding to $P = 10$, which is used for the experiments reported in the main text, is highlighted.}
    \label{ChapPPfig:nmcmc_cal_ar5}
\end{figure}

We continue our experiments without the Metropolis-Hastings (MH) correction in the kernel used within SMC, using \(\eta = 0.01\) and \(N = 150\). In Figure~\ref{ChapPPfig:nmcmc_boxplot_ar5}, we show the convergence of posterior samples as a function of the Markov chain length (\(P\)) in SMC. The discrepancy decreases uniformly as \(P\) increases; however, we do not observe significant improvements for values of \(P\) beyond \(10\), and therefore adopt this setting in the main text. Moreover, larger values of \(P\) lead to increased computational cost. We therefore choose \(P = 10\) as a trade-off between accuracy and computational efficiency.

We also report the corresponding predictive performance in Figure~\ref{ChapPPfig:nmcmc_cal_ar5}. The diagnostic curves show similar trends across training episodes, with performance improving as the number of episodes increases.

\section{Implementation details}

We present all the implementation details of our proposed algorithm in this section for the Lorenz 96 and the WeatherBench example.

\subsection{Choice of prior distribution for SMC}\label{ChapPPappendix: prior study}

\begin{figure}
    \centering
    \includegraphics[width=0.8\linewidth]{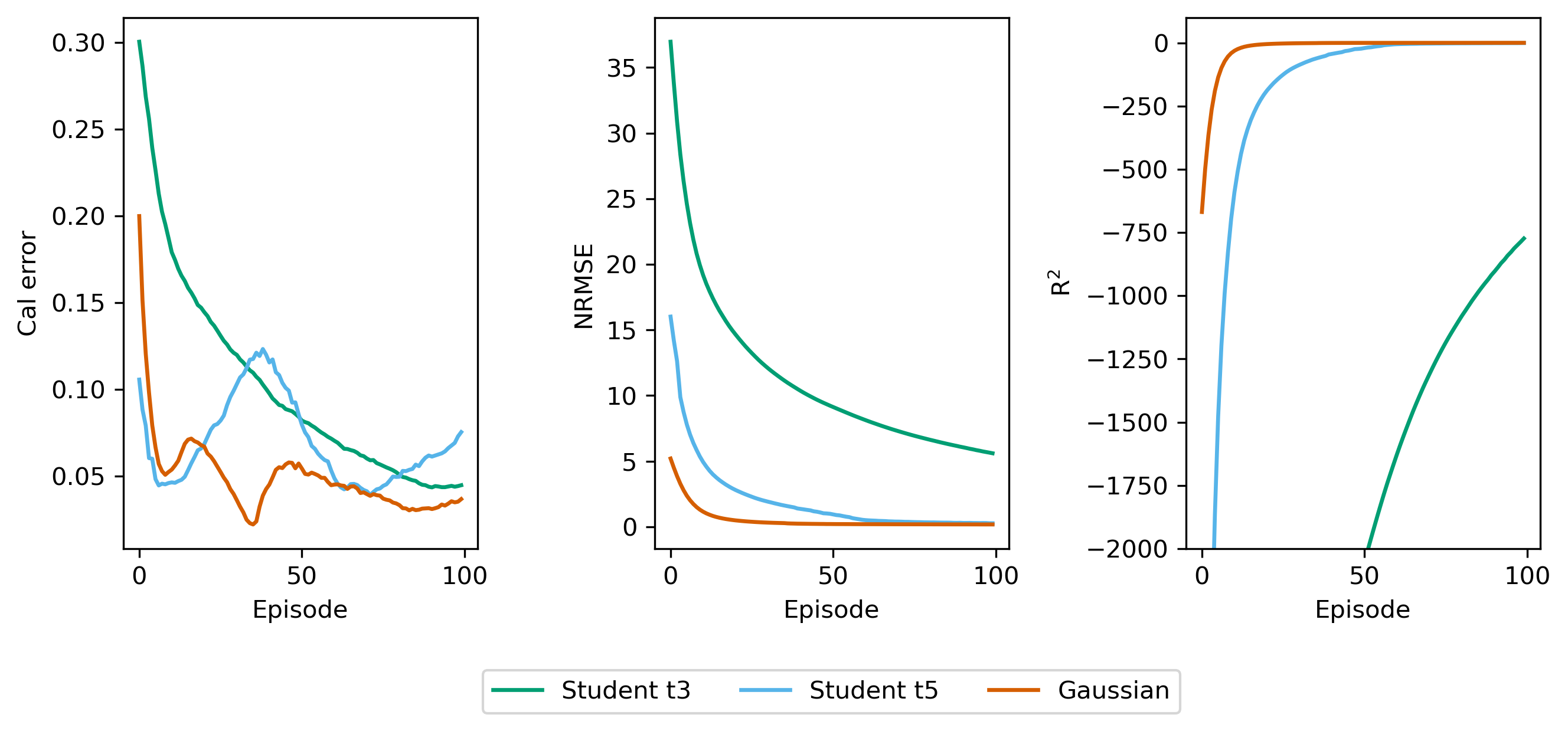}
    \caption{Predictive performance of the posterior predictives for the Lorenz~96 model under different SMC prior distributions. Calibration error, NRMSE, and \(R^2\) are shown for three priors: a standard Gaussian (orange curve), a Student's \(t\) distribution with \(3\) degrees of freedom (green curve), and a Student's \(t\) distribution with \(5\) degrees of freedom (blue curve). All priors exhibit similar overall trends, but the Gaussian prior converges noticeably faster.}
    \label{ChapPPfig:l96-prior_comparison}
\end{figure}

\begin{figure}
    \centering
    \includegraphics[width=0.8\linewidth]{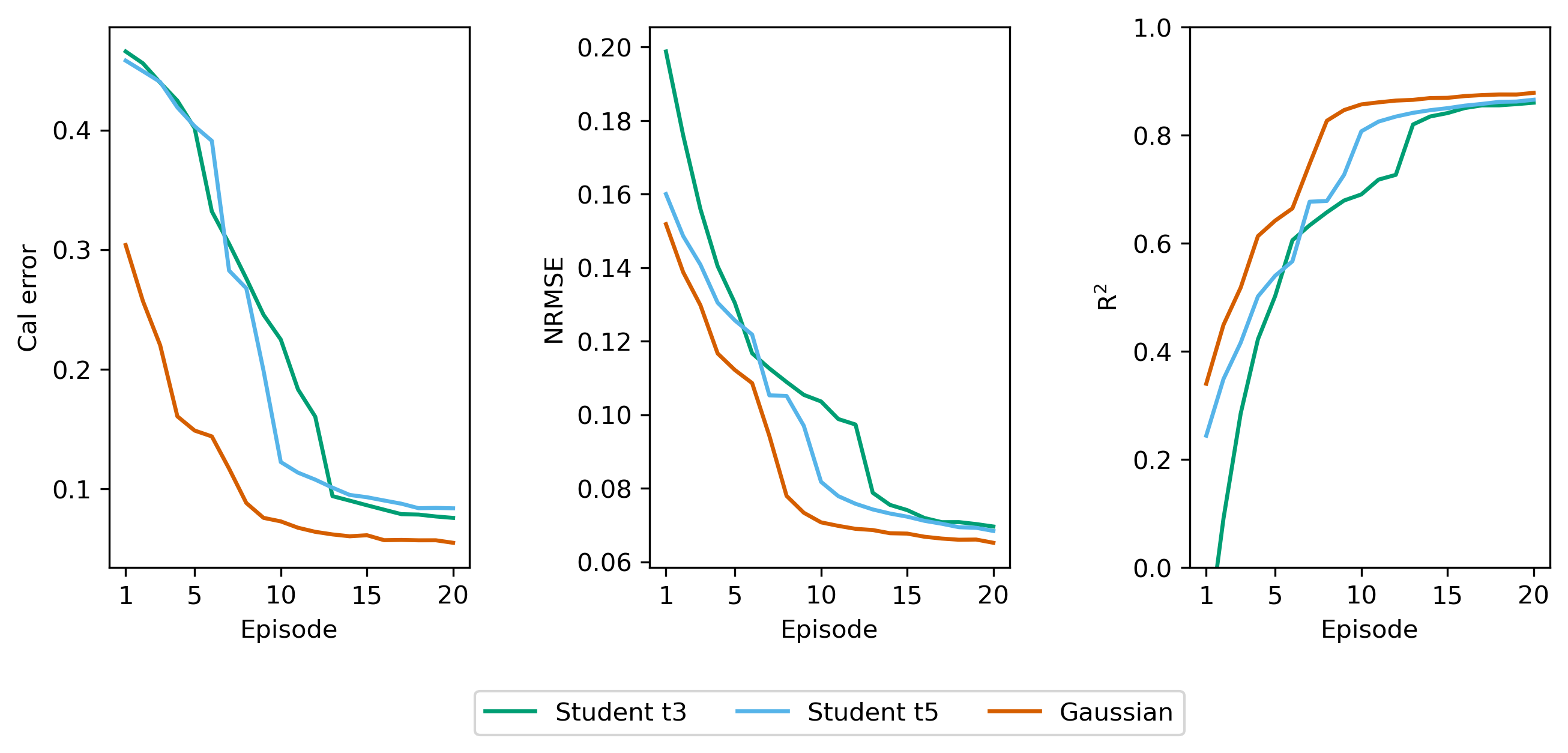}
    \caption{Predictive performance of the posterior predictives for the WeatherBench example under different SMC prior distributions. Calibration error, NRMSE, and \(R^2\) are shown for three priors: a standard Gaussian (orange curve), a Student's \(t\) distribution with \(3\) degrees of freedom (green curve), and a Student's \(t\) distribution with \(5\) degrees of freedom (blue curve). All priors show similar overall trends, but the Gaussian prior converges slightly faster.}
    \label{ChapPPfig:wb-prior_comparison}
\end{figure}

For both the Lorenz 96 and WeatherBench experiments, we set the initial proposal distribution equal to the prior distribution in the SMC procedure. We investigated the effect of different priors; specifically, a standard Gaussian and Student's t distributions with 3 and 5 degrees of freedom, on our inference results. Since the $t_1$ and $t_2$ distributions do not have a well-defined variance, we did not consider them. Figures~\ref{ChapPPfig:l96-prior_comparison} and~\ref{ChapPPfig:wb-prior_comparison} show that, although all priors exhibit similar trends in predictive accuracy across calibration error, NRMSE, and $R^2$, the Gaussian prior converges faster. Therefore, we report the Gaussian prior results in the main text.

\subsection{Details of the sampling scheme}\label{ChapPPappendix:smc table}
Here we provide the values of hyperparameters used for sampling in Table~\ref{ChapPPtab:smc-parameters}. 
\begin{table}[htbp!]
    \centering
    \begin{tabular}{|c|c|}
        \hline
        \textbf{Hyperparameter} & \textbf{Value} \\
        \hline
        Particle count in SMC, $N$ & 150\\
        MCMC chain length in SMC, $M$ & 10\\
         Learning rate, $\eta$& $10^{-6}$\\
         Kernel parameter, $\sigma_1$& $0.99$\\
         Kernel parameter, $\sigma_2$& $10^{-8}$\\
         Initial momentum, $\alpha_0$& $10^{-2}\cdot
         \mathbf{1}$\\
         CESS threshold in SMC, $\kappa$ & $0.5\cdot N = 75$\\
         \hline
    \end{tabular}
    \caption{Table containing hyperparameter values}
    \label{ChapPPtab:smc-parameters}
\end{table}

\subsection{Performance measures for probabilistic forecast}\label{ChapPPappendix-calerr}
\subsubsection{Calibration error}

Here we consider a measure of calibration of a probabilistic forecast; this measure considers the univariate marginals of the probabilistic forecast distribution $Q^\theta\left(\cdot \mid \mathbf{y}_{t-k: t-1}\right)$; for component $i$, let us denote that by $Q^{\theta, i}\left(\cdot \mid \mathbf{y}_{t-k: t-1}\right)$.

The calibration error quantifies how well the credible intervals of the probabilistic forecast $Q^{\theta, i }\left(\cdot \mid \mathbf{y}_{t-k: t-1}\right)$ match the distribution of the verification $Y_{t, i}$. Specifically, let $\alpha^{\star}(i)$ be the proportion of times the verification $y_{t, i}$ falls into an $\alpha$-credible interval of $Q^{\theta, i}\left(\cdot \mid \mathbf{y}_{t-k: t-1}\right)$, computed over all values of $t$. If the marginal forecast distribution is perfectly calibrated for component $i, \alpha^{\star}(i)=\alpha$ for all values of $\alpha \in(0,1)$.

We define therefore the calibration error as the median of $|\alpha^*(i) - \alpha|$ over 100 equally spaced values of $\alpha \in (0,1)$. Therefore, the calibration error is a value between $0$ and $1$, where
$0$ denotes perfect calibration.

\subsubsection{Normalized RMSE}

We first introduce the Root Mean-Square Error (RMSE) as
$$
\mathrm{RMSE}=\sqrt{\frac{1}{T-k} \sum_{t=k+1}^T\left(\hat{y}_{t}-y_{t}\right)^2}
$$
where we consider here for simplicity $t=k+1, \ldots, T$. From the above, we obtain the Normalized RMSE (NRMSE) as
$$
\mathrm{NRMSE}=\frac{\mathrm{R M S E}}{\max _t\left\{y_{t}\right\}-\min _t\left\{y_{t}\right\}}
$$
NRMSE $=0$ means that $\hat{y}_{t}=y_{t}$ for all $t$ 's and means a perfect fit to the observed data.

\subsubsection{Coefficient of determination}

The coefficient of determination $\mathrm{R}^2$ measures how much of the variance in $\left\{y_{t}\right\}_{t=k+1}^T$ is explained by $\left\{\hat{y}_{t}\right\}_{t=k+1}^T$. Specifically, it is given by
$$
\mathrm{R}^2=1-\frac{\sum_{t=k+1}^T\left(y_{t}-\hat{y}_{t}\right)^2}{\sum_{t=k+1}^N\left(y_{t}-\bar{y}\right)^2},
$$
where $\bar{y}=\frac{1}{T-k} \sum_{t=k+1}^T y_{t} . R^2 \leq 1$ and, when $\mathrm{R}^2=1, \hat{y}_{t}=y_{t}$ for all $t$ 's. Notice how $R^2$ is unbounded from below, and can thus be negative.

\section{Intuition for the preconditioned Kernel (Algorithm~\ref{ChapPPalg:santa-sss})}\label{ChapPPappendix:santa}

The SANTA kernel \citep{chen2016bridging} is a preconditioned, 
momentum-based SG-MCMC algorithm that can be understood as a 
Bayesian analogue of adaptive stochastic optimizers such as 
RMSprop and Adam
We summarise the key ideas here to complement the condensed 
presentation in Algorithm~1.

\paragraph{The underlying SDE.}
SANTA targets the posterior by simulating a particle $(\theta, \mathbf{p})$,
where $\theta$ are the model parameters and $\mathbf{p}$ is an 
auxiliary momentum variable, according to the SDE system 
\citep[Eq.~1]{chen2016bridging}:
\begin{align}
d\theta &= G_1(\theta)\, \mathbf{p}\, dt, \notag \\
d\mathbf{p} &= \left(-G_1(\theta)\nabla_\theta U(\theta) 
              - \Xi\, \mathbf{p} 
              + \tfrac{1}{\beta}\nabla_\theta G_1(\theta)
              + G_1(\theta)(\Xi - G_2(\theta))\nabla_\theta G_2(\theta)
              \right)dt 
              + \left(\tfrac{2}{\beta}G_2(\theta)\right)^{1/2} d\mathbf{w}, 
              \notag\\
d\Xi &= \left(\mathbf{Q} - \tfrac{1}{\beta}\mathbf{I}\right)dt,
\label{ChapPPeq:santa-sde}
\end{align}
where $\mathbf{Q} = \mathrm{diag}(\mathbf{p} \odot \mathbf{p})$, 
$\mathbf{w}$ is standard Brownian motion, $\beta$ is an inverse 
temperature, and $G_1(\theta), G_2(\theta)$ are Riemannian metrics 
approximated in practice by the RMSprop preconditioner 
$\mathbf{g}_t = \mathbf{1} \oslash \!\sqrt{\lambda + \sqrt{\mathbf{v}_t}}$,
with the running gradient-variance estimate
$\mathbf{v}_t = \sigma\, \mathbf{v}_{t-1} + \frac{1-\sigma}{N^2}\,
\tilde{\mathbf{f}}_t \odot \tilde{\mathbf{f}}_t$. Intuitively, the preconditioner rescales 
the gradient and noise by the inverse square root of a running 
estimate of the per-coordinate gradient variance, so that 
coordinates with large, noisy gradients take smaller steps. 
The thermostat variable $\Xi$ acts as an adaptive friction: 
it damps the momentum to prevent particles from overshooting, 
while the injected Brownian noise maintains posterior diversity.
We keep the inverse temperature fixed at $\beta=1$.

\paragraph{Symmetric splitting discretisation.}
Directly simulating \eqref{ChapPPeq:santa-sde} is infeasible, so 
\citet{chen2016bridging} split it into three analytically 
tractable sub-SDEs and solve them in the order A--B--O--B--A 
with half-steps $h/2$ on A and B and a full step $h$ on O.
The analytic solutions for stepsize $h$ are 
\citep[Supplementary A, Eq.~6]{chen2016bridging}:
\begin{align*}
\textbf{A:}&\quad 
  \theta_t = \theta_{t-1} + G_1(\theta)\,\mathbf{p}\,h,\quad
  \Xi_t = \Xi_{t-1} + \left(\mathbf{Q} - \tfrac{1}{\beta}\mathbf{I}\right)h,\\
\textbf{B:}&\quad 
  \mathbf{p}_t = \exp(-\Xi\, h)\,\mathbf{p}_{t-1},\\
\textbf{O:}&\quad 
  \mathbf{p}_t = \mathbf{p}_{t-1} 
  + \!\left(-G_1(\theta)\nabla_\theta U(\theta) 
    + \tfrac{1}{\beta}\nabla_\theta G_1(\theta)
    + G_1(\theta)(\Xi - G_2(\theta))\nabla_\theta G_2(\theta)
  \right)h
  + \left(\tfrac{2}{\beta}G_2(\theta)\right)^{1/2}\!\zeta_t,
\end{align*}
where $\zeta_t \sim \mathcal{N}(\mathbf{0}, \mathbf{I})$.
Substituting the RMSprop preconditioner for $G_1$ and $G_2$ 
yields Algorithm~2 of \citet{chen2016bridging}, which is the 
kernel used in our Algorithm~1. The A--B--O--B--A ordering is 
analogous to the leapfrog integrator in Hamiltonian Monte Carlo 
\citep{neal2011mcmc}, and gives a second-order accurate 
discretisation with provably controlled bias and MSE 
\citep[Theorem~2]{chen2016bridging}.

\end{document}